\pdfoutput=1
\documentclass[11pt]{article}
\usepackage[final]{acl}
\usepackage{times}
\usepackage{latexsym}
\usepackage[T1]{fontenc}
\usepackage[utf8]{inputenc}
\usepackage{microtype}
\usepackage{inconsolata}
\usepackage{booktabs}
\usepackage{microtype}
\usepackage{enumerate}
\usepackage{enumitem}
\setenumerate[1]{itemsep=0pt, partopsep=0pt, parsep=\parskip, topsep=2pt}
\setitemize[1]{itemsep=0pt, partopsep=0pt, parsep=\parskip, topsep=2pt}
\setdescription{itemsep=0pt, partopsep=0pt, parsep=\parskip, topsep=2pt}
\usepackage{booktabs}
\usepackage{colortbl}
\usepackage{color}
\usepackage{ctable}
\usepackage{microtype}
\usepackage{amsfonts,amssymb}
\usepackage{amsmath}
\usepackage{multirow}
\usepackage{bbm}
\usepackage{booktabs}
\usepackage{graphicx}
\usepackage{color}
\usepackage{colortbl}
\usepackage{float}
\usepackage{framed}
\usepackage{wrapfig}
\usepackage{tcolorbox}
\usepackage{array}
\usepackage{xcolor}
\usepackage{caption}
\usepackage{inconsolata}
\usepackage{algorithm}
\usepackage{algpseudocode}
\usepackage{makecell}
\usepackage{ulem}
\usepackage{titlesec}
\usepackage{textcomp}
\usepackage{marvosym}
\usepackage{fontawesome}

\title{\raisebox{-0.8ex}{\includegraphics[scale=0.12]{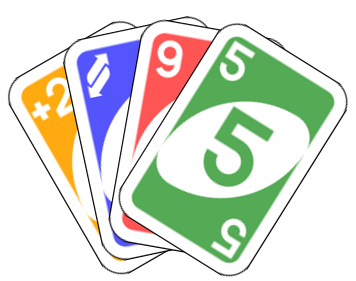}}\hspace{0.2cm}UNO Arena for Evaluating Sequential Decision-Making Capability of Large Language Models}

\author{Zhanyue Qin$^{1}$\textsuperscript{\Letter}, Haochuan Wang$^{1}$, Deyuan Liu$^{1}$, Ziyang Song$^{1}$, Cunhang Fan$^{2}$, Zhao Lv$^{2}$, \\ \bf{Jinlin Wu$^{3}$, Zhen Lei$^{3}$, Zhiying Tu$^{1}$, Dianhui Chu$^{1}$, Xiaoyan Yu$^{3}$, Dianbo Sui$^{1}$\textsuperscript{\Letter}
\thanks{Dianbo Sui is the corresponding author.}} \\
$^{1}$ Harbin Institute of Technology\\ 
$^{2}$ Anhui University\\ $^{3}$ Institute of Automation, Chinese Academy of Sciences\\
johnneyqin@gmail.com, suidianbo@hit.edu.cn }

\begin{document}
\maketitle
\begin{abstract}
Sequential decision-making refers to algorithms that take into account the dynamics of the environment, where early decisions affect subsequent decisions. With large language models (LLMs) demonstrating powerful capability between tasks, we can't help but ask: \textit{Can Current LLMs Effectively Make Sequential Decisions?} In order to answer this question, we propose the UNO Arena based on the card game UNO to evaluate the sequential decision-making capability of LLMs and explain in detail why we choose UNO. In UNO Arena, We evaluate the sequential decision-making capability of LLMs dynamically with novel metrics based Monte Carlo methods. We set up random players, DQN-based reinforcement learning players, and LLM players (e.g. GPT-4, Gemini-pro) for comparison testing. Furthermore, in order to improve the sequential decision-making capability of LLMs, we propose the \textbf{\textsc{TuTri}} player, which can involves having LLMs reflect their own actions wtih the summary of game history and the game strategy. Numerous experiments demonstrate that the \textbf{\textsc{TuTri}} player achieves a notable breakthrough in the performance of sequential decision-making compared to the vanilla LLM player.
\end{abstract}

\section{Introduction}
In artificial intelligence, sequential decision-making refers to algorithms that take the dynamics of the world into consideration~\cite{frankish2014cambridge}, and it can be described as a procedural approach to decision-making, or as a step by step decision theory. Sequential decision-making has as a consequence the intertemporal choice problem, where earlier decisions influences the later available choices~\cite{amir2014reasoning}.

In recent years, Large language models (LLMs) are gaining increasing popularity in both academia and industry, owing to their unprecedented performances in various applications~\cite{chang2023survey}, ranging from chatbots to medical diagnoses~\cite{wang2023chatcad} to robotics~\cite{he2022controlling}. From robots handle complex tasks~\cite{amiri2020learning} to entrepreneurial action~\cite{mcmullen2015entrepreneurial}, sequential decision-making permeates diverse domains. Hence, an interesting question arises: \textit{Can Current LLMs Effectively Make Sequential Decisions?}

To answer this question, we need to design a benckmark to evaluate the sequential decision-making ability of LLMs. However, evaluating LLMs' abilities is not trivial. Many works have been proposed to test LLMs' performances on either a large-scale static benchmark such as MMLU~\cite{hendrycks2021measuring}, or with A/B tests judged by humans~\citep{ganguli2023challenges}.
One common and evident limitation of these methods, however, is that the environment for LLMs to be tested is static~\citep{aiyappa2023trust,zhou2023dont}, which can not reflect the domino effect in sequential decision-making. Besides, data contamination~\cite{sainz-etal-2023-nlp,zeng2024mrgsm8k,xu2024benchmark}, which means the inclusion of test data examples and labels in the pre-training data, also challenges the efficacy of these static benchmarks in differentiating model capabilities. 

\begin{figure*}[htbp]
\centering
\includegraphics[width=1\textwidth]{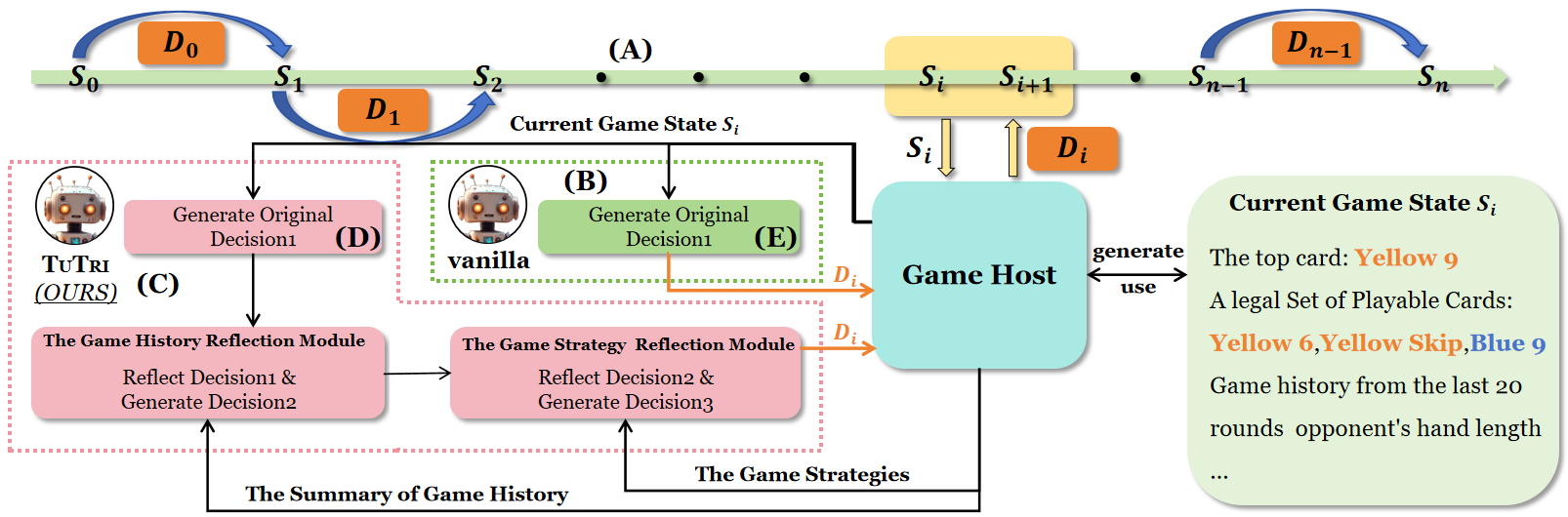}
\caption{In this figure, (A) demonstrates the sequential decision-making process in UNO Arena, (B) shows the execution process of the vanilla LLM player, and (C) shows the execution process of the \textsc{TuTri} player. In fact, The Module (D) and the Module (E) are completely identical.}
\label{UNO_figure}
\end{figure*}

Unlike static evaluation, dynamic evaluation by treating LLMs as game-playing agents attracted more and more attention of researchers recently, such as beauty contests and private-value second price auctions~\cite{guo2024economics}, Warewolf~\cite{xu2023exploring}, Avalon~\cite{wang2023avalons,light2023avalonbench}, Leduc Hold’em~\cite{guo2023suspicionagent}. However, current attempts do not account for sequential decision-making, and these games are either challenging to evaluate for intermediate results (such as Werewolf) or have too few decision points per round (such as Leduc Hold’em). Meanwhile, we should also note that studies of dynamically evaluating sequential decision-making capability  in reinforcement learning, such as games like Go~\cite{silver2017mastering}, Dou Di Zhu~\cite{you2019combinational}, and Mahjong~\cite{li2020suphx}. However, these games present an excessively large action space. For instance, in Dou Di Zhu, players can use any combination of their cards each round, posing significant challenges for current LLMs~\cite{zhai2024finetuning}.

Considering the above aspects, we make the following efforts in this paper: 

First, we build UNO Arena to dynamically evaluate the sequential decision-making capability of current LLMs. In UNO arena, we allow LLMs to participate as players in the UNO game\footnote{\url{https://en.wikipedia.org/wiki/Uno_(card_game)}}, aiming to play all the cards in their hand as quickly as possible. Compared to games like Leduc Hold'em, which have fewer moves per game, UNO features an average of dozens of moves per game, making it an ideal testbed for sequential decision-making~\cite{TacklUNO2021}. Additionally, unlike common games in reinforcement learning, legal actions in the UNO Arena are limited, only including drawing cards, playing cards, selecting colors to convert, and choosing whether to challenge the wild draw four card. Furthermore, to monitor the  behaviours of LLMs in the UNO arena, we propose some real-time quantitative evaluation metrics by leveraging the Monte Carlo method~\cite{kroese2014monte} as the reference, which provide a window to observe the intermediate results and various phenomenons (like domino effect) in LLMs' sequential decision-making. 

Second,  based on the proposed UNO Arena, we set up a family of strong and representative players. In detail, we first build the random player, which makes decisions based on chance rather than a specific, consistent plan, without considering the game's current state or potential outcomes. Then, we implement the reinforcement learning based player, which leverages DQN~\cite{mnih2013playing}
to develop sophisticated strategies for playing UNO. Finally, to probing the capability of LLMs in sequential decision-making, we provides the task description and then prompt LLMs, like  GPT-4~\cite{achiam2023gpt} and Gemini-pro~\cite{team2023gemini}, to generate their reasoning steps that lead to the final action. 
 
Third, to  unleash the  fully potential capability of LLMs in sequential decision-making, we propose the \textbf{\textsc{TuTri}} player with reflection mechanism~\cite{shinn2024reflexion}, which can involves having LLMs analyze their own actions wtih the game history and the game strategy. In detail, the proposed agent framework consists of two key reflection modules: the game history reflection module and the game strategy reflection module. In the game history reflection module, we provide the statistical data of game history and then prompt the LLMs to rethink their decision, which simulates the process of card memorization by humans when playing UNO. In the game strategy reflection module, LLMs further take into account the game strategy, like saving wild draw four, and proceed to make the final decision, which simulates the use and adherence to strategies by humans when playing UNO. 

In the experiment, we comprehensively evaluate some mainstream LLMs' ability of sequential decision-making, including GPT-3.5~\cite{ChatGPT}, GPT-4~\cite{achiam2023gpt}, Gemini-pro~\cite{team2023gemini}, Llama 2~\cite{touvron2023llama}, and ChatGLM3~\cite{du2021glm, zeng2022glm}. Our experiments show that among these LLMs, GPT-4 is the most effective sequential decision-maker.

In summary, our contributions are as follows:
\begin{itemize}
    \item We propose a dynamic evaluation method named UNO Arena for assessing the sequential decision-making capability of large language models (LLMs) based on the card game UNO. This method supports the evaluation of 2-10 LLM players, reinforcement learning players, or random players engaged in a single UNO game.
    \item We introduce multiple unique evaluation metrics based on the Monte Carlo method for evaluating the sequential decision-making capabilities of players in UNO Arena.
    \item To improve the sequential decision-making capabilities of LLMs and enhance their performance in the highly dynamic and complex UNO game, we have developed the \textbf{\textsc{TuTri}} player and compared it horizontally with the vanilla LLM player.
\end{itemize}

\section{Related Work}

\textbf{Evaluate LLMs Dynamically with Game}: LLMs has presented increasingly emerging ability on game-playing~\cite{brookins2023playing, akata2023playing} in recent development and iterations. 
\citet{wang2023avalons} use the Avalon, which contains elements of deception, to evaluate the capability of LLMs to recognize and handle deceptive information.
\citet{gong2023mindagent} leverage the CuisineWorld and Minecraft to assess the planning and emergency cooperation capabilities of LLMs.
\citet{guo2024economics} employ beauty contests and auction games to evaluate the rationality, strategic reasoning capability, and adherence to instructions of LLMs.
\citet{xu2023exploring} use the game Werewolf to evaluate the capability of LLMs to infer player roles. Despite evaluating with game becoming a popular trend, exploring into sequential decision-making capability is still of scarcity in current works.
\\[3pt]
\textbf{Development of Agent Framework}: LLM agents have been perceived as a promising way to realizing Artificial General Intelligence(AGI)~\cite{xi2023rise} and recently have shown emergent abilities to execute various tasks in complex environment~\cite{wei2022emergent}. SiLLM~\cite{guo2024sillm} merges large language models with synchronous machine translation, using policy decision agents and translation agents. 
LLM-Vectorizer~\cite{taneja2024llm} uses multiple agents to generate vectorized code by leveraging large language models and test-based feedback. We tailored a special framework for UNO, featuring self-refinement and iterative thinking.
\\[3pt]
\textbf{Sequential Decision-Making Capability}: Sequential decision-making refers to the process of making a series of decisions over time, where each decision may impact future choices and outcomes~\cite{amir2014reasoning}. Though certain algorithms or reinforce learning provide solutions for some sequential decision-making problems~\cite{littman1996algorithms}, LLM-based sequential decision-making are only employed in limited field like recommendation~\cite{wang2023drdt}. In our work, we utilized UNO, which is not an easy one even for human~\cite{demaine2014uno}, to explore the sequential decision-making ability of LLMs. With certain methods like integrating past experiences and expert advice or demonstrations~\cite{chen2023introspective}, we made efforts to maximally leverage the decision making ability as possible in a sequential manner.

\section{The UNO Arena} 
In this section, we first provide a brief overview of the version of UNO we adopt in the subsection §\ref{subsection: 3.1}. Then, we present the four different types of players in the UNO arena in the subsection §\ref{subsection: 3.2}. Next, we detail how to use Monte Carlo methods to determine whether a player has made an optimal decision in subsection §\ref{subsection: 3.3}. In the end, we introduce our evaluation metrics in subsection §\ref{subsection: 3.4}.

\subsection{The UNO Game}
\label{subsection: 3.1}
We select the UNO as the foundation within our arena due to its widespread popularity, simplicity and mathematical value. There are various versions of the UNO game. In this section, we briefly introduce the rules of the version we adopt in this work.

\textbf{UNO Cards:} A deck of UNO cards comprises a total of 108 cards. UNO cards are divided into three types: number cards, function cards, and wild cards. A number card is composed of a color (Red, Blue, Yellow and Green) and a number (ranging from 0 to 9). A function card is composed of a color (Red, Blue, Yellow and Green) and a function (Skip, Reverse, Draw Two). The wild cards has no color and is only composed of Wild cards and Wild Draw Four cards. The effects of the function cards and wild cards are shown in the Table~\ref{tab:func_card}.

\textbf{UNO Process:} First, deal each player 7 initial cards in clockwise order, then continue drawing cards until a number card is drawn and set as the top card of the initial discard pile. All players take rounds playing cards in clockwise order(it will be reversed by a reverse card) until a player runs out of his cards or the draw pile is exhausted, signaling the end of the game.

\textbf{UNO Action:} From the beginning to the end of the game, players continuously take actions in UNO. In our work, UNO includes the following types of actions:
\begin{itemize}
    \item \textbf{Select Card}: When a player comes his playing round, they need to play a card that matches the color, number, or function of the top card in the discard pile, or play a Wild card. If they don't have a card to play, they must draw one card.
    \item \textbf{Select Color}: After a player plays a Wild card or a Wild Draw Four card, they need to change the color of the current top card to one of Red, Yellow, Blue or Green.
    \item \textbf{Select ChallengeFlag}: After a player's previous opponent plays a Wild Draw Four card, the player needs to decide whether to challenge the legality of the previous opponent's Wild Draw Four card.
\end{itemize}

For more details about the UNO games, please refer to Appendix \ref{UNO_appendix}. The Figure \ref{UNO_figure} (A) shows the workflow diagrams of UNO Arena.

\subsection{Players in the UNO Arena}
\label{subsection: 3.2}
In the UNO Arena, we initially involve three types of players: 
random player, reinforcement learning based player, vanilla LLM player. To further unleash the potential capability of
LLMs in sequential decision-making,  we propose \textsc{TuTri} player, which involves reflection mechanism. 

\begin{table}[t]
    \centering
    \scriptsize
    \begin{tabular}{>{\raggedright\arraybackslash}p{25pt}|>{\raggedright\arraybackslash}p{25pt}|>{\raggedright\arraybackslash}p{135pt}}
        \toprule
        \makecell{\textbf{Card}} & \makecell{\textbf{Sample}} & \makecell{\textbf{Effect}} \\
        \midrule
        \midrule
        \makecell{Skip} & \makecell{\includegraphics[width=0.05\textwidth]{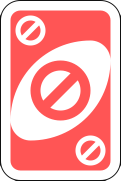}}            & \makecell{The next player in sequence misses a round.} \\
        \midrule
        
        \makecell{Reverse} & \makecell{\includegraphics[width=0.05\textwidth]{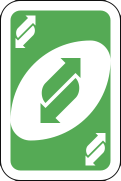}}         & \makecell{Order of play switches directions (clockwise to\\counterclockwise, or vice versa).} \\
        \midrule
        
        \makecell{Draw\\Two} & \makecell{\includegraphics[width=0.05\textwidth]{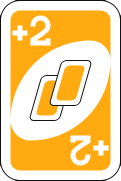}}        & \makecell{The next player in sequence draws two cards\\and misses a round.} \\
        \midrule
        
        \makecell{Wild} & \makecell{\includegraphics[width=0.05\textwidth]{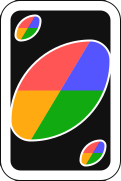}}            & \makecell{Player declares the next color to be matched\\(it can be used on any round even if the player\\has any card of matching color).} \\
        \midrule
        
        \makecell{Wild\\Draw\\Four} & \makecell{\includegraphics[width=0.05\textwidth]{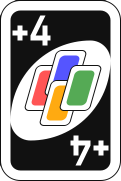}}  & \makecell{Player declares the next color to be matched.\\The next player in sequence draws four cards\\and misses a round. May be legally played if\\the player has cards of the current color.} \\
        \bottomrule
    \end{tabular}
    \caption{The effects of function and wild cards.}
    \label{tab:func_card}
\vspace{-20pt}
\end{table}

\textbf{Random Player}: As like its name suggests, the random player performs all actions randomly, such as randomly selecting a regulative card to play when it's their turn. The random player can be considered the baseline of the UNO Arena, mainly serving to maintain the flow of the UNO game. If some players outperform the random player, we can infer that these players are consciously playing UNO with an understanding of the game rules.

\textbf{Reinforcement Learning Based Player}: Previous research has sought breakthroughs in UNO using reinforcement learning models~\cite{TacklUNO2021}. We built our reinforcement learning player with DQN~\cite{mnih2013playing} model based on the open-source project RLcard~\cite{zha2019rlcard}.

\textbf{Vanilla LLM Player}:
During the vanilla LLM player's turn, the game host transmits all publicly available information through a prompt to the LLM. The LLM then returns a JSON containing the decision and reasoning as required by the prompt. The Figure \ref{UNO_figure} (B) shows the workflow diagrams of vanilla LLM player.

\textbf{\textsc{TuTri} Player}: While LLMs do not 
always generate the best output on their first try just as human~\cite{madaan2023self}, iterative feedback and refinement could be a necessity for a better agent framework. Moreover, human-like thinking patterns, such as introspective reflections foster divergent thinking processes~\cite{zhang2023exploring}, inspires us to propose the \textsc{TuTri} player. This advanced framework is designed to navigate the intricacies of UNO game play, offering a more structured approach to strategic sequential decision-making. The original decision for \textsc{TuTri} player is exactly the same as the vanilla LLM player's decision, after that are two additional reflection modules.

\begin{itemize}
    \item \textbf{The Game History Reflection Module}: In the module, we provide statistical information about game history to \textsc{TuTri} player, and they are told to \textit{reflect the action you just selected} with these auxiliary information. Just like human thinking when playing UNO, if there is a large number of green cards that have already been played in the game's history, it is very advantageous for players to play a green card. LLM should output both reflection thoughts and updated action.
    \item \textbf{The Game Strategy Reflection Module}: In the module, we provide additional useful game strategies to \textsc{TuTri} players, and they are again told to \textit{reflect the action you just selected} based on game strategies. For example, since wild cards can be played at any situations and disrupt other players, saving the wild cards in your hand as long as possible is a very useful game strategy. LLM should output both reflection thoughts and updated action (the final action).
\end{itemize}

It must be emphasized that the \textsc{TuTri} player should work in a conversational manner, with exactly 3 times Q\&A per round. Moveover, the \textsc{TuTri} players may keep their original decision, in other words, literally updating the action is not a necessity, nevertheless, the reflection process, instead of simple I-O prompting of interaction, providing more opportunities for mistake correcting and divergent thinking. The Figure \ref{UNO_figure}(C) shows the workflow diagrams of \textsc{TuTri} player.
\subsection{Monte Carlo Simulation Method for Monitoring  Players' Behavior}
\label{subsection: 3.3}
In the game play, the change in each player's winning rate after making a decision is the key for tracking.  In the classical combinatorial games, like Nim~\cite{bouton1901nim} or Wythoff's Game~\cite{wythoff1907modification}, positions space are limited and thus computationally affordable, while UNO is more intricate, where positional space exponentially increases as cards number increases and the calculation gets tougher~\cite{demaine2014uno}.

To make a plausible ranking mechanism of the candidate decisions, we define the concept of \textbf{optimal decision}, meaning the state transferred by the decision from last state, has a highest winning rate concerning all subsequent outcomes, and thus adopt Monte Carlo Simulation~\cite{mooney1997monte} to calculate the estimated winning rate.

Detailedly, with $S_i$ representing the state of the game after the $i$-th step taken, $D_{i,j}$ representing the $j$-th legal decision candidates at state $S_i$, $\mathcal{T}$ representing the state transfer function, $\mathcal{E}$ representing the estimate function of state, thereby we have the definition of the optimal decision $D_{i, opt}$ at the $i$-th step where
\begin{equation}
    opt = \mathop{\arg \max}_{j} \mathcal{E}(\mathcal{T}(S_{i-1}, D_{i, j}))
\end{equation}

In calculation of $\mathcal{E}(S_i)$, we massively randomly generate the subsequent decision sequence $\{D_{i+1}, D_{i+2}, \dots\}$ and thus obtain the subsequent state sequence $\{S_{i+1}, S_{i+2}, \dots\}$. Then $\mathcal{E}(S_i)$ is assigned to the ratio of number of sequences where the player plays the state $S_{i-1}$ comes as the winner, to the total number of sequences simulated.

As the times we simulate the subsequent sequence increases, the approximate value $\mathcal{E}(S_i)$ gets more precise, though we could not enumerate all the possible situations. To balance the time expenditure and the precision of the metrics, we control the simulation times in a certain range. Additionally, a threshold parameter $p$ is set to identify critical decisions. We say a decision $D:= D_i$ is \textbf{critical} if among its all candidate choices $D_j$
\begin{equation}
\max \mathcal{E}(\mathcal{T}(S, D_j)) - \min \mathcal{E}(\mathcal{T}(S, D_j)) \geq p
\end{equation}
Actual decisions made on critical positions may have a huge effect on the winning rate, which is consistent with the game nature.

\subsection{Evaluation Metrics in the UNO Arena} 
\label{subsection: 3.4}
In our work, we design three evaluation metrics, including WR, ODHR@K and  ADR@K,  in conjunction with the UNO  game to comprehensively evaluate the sequential decision-making capability of LLMs. Among these metrics, ODHR@K and  ADR@K can off a glimpse into the intermediate results  in the sequential decision-making of LLMs.
\\[3pt]
\textbf{Winning Rate (WR)}. WR denotes the proportion of player wins to total game innings, and can be represented as:
\begin{equation}
        \text{WR} = \frac{N_{\text{Win}}}{N_\text{{Game}}}
\end{equation}
where $N_{\text{Win}}$ represents the total number of times the given player has won, and $N_\text{{Game}}$ denotes the total game innings.
\\[3pt]
\textbf{Optimal Decision Hit Rate at K Decision Points (ODHR@K)} : This metric measure the proportion of times players make the best decision to all decision times, when facing K decision points:
    \begin{equation}
        \text{ODHR@K} = \frac{\text{$N_{\text{Hit@K}}$}}{\text{$N_{\text{Decision@K}}$}}
    \end{equation}
  where $N_{\text{Hit@K}}$ is the number of times the player makes the optimal decision when facing K optional decision points, and $N_{\text{Decision@K}}$ represents the total number of times the agent player makes decision when it faces K optional decision points. 
\\[3pt]
 \textbf{Average Decision Rank at K Decision Points (ADR@K)}. This metric looks at the rank of output decision made by the player, and can be denoted as:
    \begin{equation}
    \text{ADR@K} = \frac{\sum_{i=1}^{\text{$N_{\text{Decision@K}}$}}\text{Rank}(D_i)}{\text{$N_{\text{Decision@K}}$}}
    \end{equation}
  where $\text{Rank}(D_i)$ represents the rank from best to worst among all legal decisions in its decision-making process, and $N_{\text{Decision@K}}$ represents the total number of times the agent player makes decision when it faces K optional decision points. 

For metrics ODHR@K and ADR@K, according to the characteristics of UNO, we only focus on the situations where $K$ is equal to $2$, $3$ or $4$, because the vast majority of decisions in UNO do not exceed 4~\cite{TacklUNO2021}.


\section{Experiments}
In this section, we first conduct preliminary experiments with vanilla LLM players, RL players, and random players in subsection §\ref{subsection: 4.1}. Then, we have multiple different LLM-based vanilla players compete in UNO Arena to identify the best LLM in subsection §\ref{subsection: 4.2}. Next, we test the superiority of the \textsc{TuTri} players compared to the vanilla LLM players in subsection §\ref{subsection: 4.3}. Finally, we perform ablation experiments on the \textsc{TuTri} player in subsection §\ref{subsection: 4.4}.

To ensure the generalization of the experiments, we take the mainstream LLMs mentioned in the introduction: (1) gpt-3.5-turbo-16k-0613~\cite{ChatGPT}; (2) gpt-4-1106-preview~\cite{achiam2023gpt}; (3) Gemini-pro~\cite{team2023gemini}; (4) Llama-2-7b-chat~\cite{touvron2023llama}; (5) ChatGLM3-6b~\cite{du2021glm, zeng2022glm}.

\subsection{1v1 UNO Arena between vanilla LLM players, RL players and random players}
\label{subsection: 4.1}
In order to verify the rationality of using UNO Arena to evaluate the sequential decision-making ability of LLMs, we first conduct experiments on vanilla LLM players, RL players and random players in 1V1 UNO Arena. We randomly generate 500 sets of UNO initial decks. Each vanilla LLM player or RL player have to play with the random player in these 500 initial decks. In addition, the random players are the first to play cards in all games. The results are shown in Table \ref{exp_result_table_1}. 

\begin{table}[ht]
\centering
\scalebox{0.6}{
\begin{tabular}{@{}r|cccccc@{}}
\toprule
\multirow{2}{*}{Metrics} & \multicolumn{6}{c}{Vanilla LLM Players \& RL Player with DNQ} \\
\cmidrule(lr){2-7}
& \multicolumn{1}{c}{GPT-3.5} & \multicolumn{1}{c}{GPT-4} &  \multicolumn{1}{c}{Gemini-Pro} & \multicolumn{1}{c}{Llama 2} & \multicolumn{1}{c}{ChatGLM3} & \multicolumn{1}{c}{DNQ} \\
\midrule[\heavyrulewidth]
WR ($\uparrow$) & 55.80 & \textbf{63.20} & 53.80 & 53.60 & 48.80 & \uline{57.40} \\
\midrule
ODHR@2 ($\uparrow$) & 57.34 & \textbf{61.47} & 53.94 & 53.69 & 49.75 & \uline{54.96} \\
DAR@2 ($\downarrow$) & 1.427 & \textbf{1.385} & 1.461 & 1.463 & 1.503 & \uline{1.450} \\
\midrule
ODHR@3 ($\uparrow$) & 32.15 & \textbf{39.30} & 34.42 & 33.84 & 34.45 & \uline{35.98} \\
DAR@3 ($\downarrow$) & 2.010 & \textbf{1.904} & 2.017 & 1.994 & 2.034 & \uline{1.947} \\
\midrule
ODHR@4 ($\uparrow$) & 27.20 & \textbf{36.99} & 31.05 & 27.39 & 25.36 & \uline{37.74} \\
DAR@4 ($\downarrow$) & 2.399 & \textbf{2.142} & 2.331 & 2.436 & 2.460 & \uline{2.247} \\
\bottomrule
\end{tabular}}
\caption{Statistical results of random player VS vanilla LLM player or RL player with DNQ. The decision threshold $p$ for critical decision in ODHR@K and ADR@K is 0.15. Bold indicates the best result, underline the second best result, and the Table \ref{exp_result_table_2} below follows this pattern.}
\label{exp_result_table_1}
\end{table}

\begin{table}[ht]
\begin{center}
\scalebox{0.67}{
\begin{tabular}{@{}r|ccccc@{}}
\toprule
\multirow{2}{*}{Metrics} & \multicolumn{5}{c}{Vanilla LLM Players}\\
\cmidrule(lr){2-6}
\empty & GPT-3.5 & GPT-4 & Gemini-Pro & Llama 2 & ChatGLM3\\
\midrule[\heavyrulewidth]
WR ($\uparrow$)&\uline{22.80}&\textbf{24.20}&20.40&20.00&15.60 \\
\midrule
ODHR@2 ($\uparrow$)&52.57&\textbf{54.77}&49.88&\uline{54.08}&50.52\\
DAR@2 ($\downarrow$)&1.474&\textbf{1.452}&1.501&\uline{1.459}&1.495\\
\midrule
ODHR@3 ($\uparrow$)&\uline{39.56}&\textbf{41.41}&33.14&34.78&33.13\\
DAR@3 ($\downarrow$)&\uline{1.889}&\textbf{1.885}&2.034&1.978&2.043\\
\midrule
ODHR@4 ($\uparrow$)&\uline{26.75}&\textbf{29.03}&25.74&24.90&25.04\\
DAR@4 ($\downarrow$)&\uline{2.407}&\textbf{2.366}&2.516&2.471&2.477\\
\bottomrule
\end{tabular}}
\caption{Statistical results of competition among 5 vanilla LLM players in UNO Arena. The decision threshold $p$ for critical decision in ODHR@K and ADR@K is 0.00.}
\label{exp_result_table_2}
\end{center}
\end{table}

\begin{table*}[ht]
\begin{center}
\scalebox{0.63}{
\begin{tabular}{@{}l|l|ll|ll|ll@{}}
\toprule
LLM
& WR ($\uparrow$)& ODHR@2 ($\uparrow$)& DAR@2 ($\downarrow$)& ODHR@3 ($\uparrow$)& DAR@3 ($\downarrow$)& ODHR@4 ($\uparrow$)& DAR@4 ($\downarrow$)\\
\midrule[\heavyrulewidth]
GPT-3.5 (vanilla) &
48.00 & 53.05 & 1.4695 & 34.97 & 1.9508 & 34.47 & 2.2340\\
GPT-3.5 (\textsc{TuTri}) &
52.50~\textcolor{red}{(+4.50\%)} & 54.01~\textcolor{red}{(+0.06\%)} & 1.4599~\textcolor{red}{(-0.06\%)} & 43.13~\textcolor{red}{(+8.16\%)} & 1.8563~\textcolor{red}{(-4.73\%)} & 32.92~\textcolor{blue}{(-1.55\%)} & 2.2667~\textcolor{blue}{(+1.09\%)}\\
\midrule
GPT-4 (vanilla) &
49.00 & 56.27 & 1.4373 & 39.38 & 1.9375 & 36.24 & 2.2140\\
GPT-4 (\textsc{TuTri}) &
51.00~\textcolor{red}{(+2.00\%)} & 56.60~\textcolor{red}{(+0.33\%)} & 1.4340~\textcolor{red}{(-0.33\%)} & 40.14~\textcolor{red}{(+0.76\%)} & 1.8592~\textcolor{red}{(-3.92\%)} & 36.33~\textcolor{red}{(+0.09\%)} & 2.1510~\textcolor{red}{(-2.10\%)}\\
\midrule
Gemini-pro (vanilla) &
44.00 & 50.62 & 1.4938 & 37.04 & 2.0159 & 25.44 & 2.4737\\
Gemini-pro (\textsc{TuTri}) &
56.50~\textcolor{red}{(+12.50\%)} & 53.64~\textcolor{red}{(+3.02\%)} & 1.4636~\textcolor{red}{(-3.02\%)} & 34.13~\textcolor{blue}{(-2.91\%)} & 1.9461~\textcolor{red}{(-3.49\%)} & 30.36~\textcolor{red}{(+4.92\%)} & 2.3482~\textcolor{red}{(-4.18\%)}\\
\midrule
Llama 2 (vanilla) &
47.00 & 49.54 & 1.5046 & 33.11 & 1.9595 & 29.11 & 2.3944\\
Llama 2 (\textsc{TuTri}) &
54.00~\textcolor{red}{(+7.00\%)} & 55.07~\textcolor{red}{(+5.53\%)} & 1.4493~\textcolor{red}{(-5.53\%)} & 37.31~\textcolor{red}{(+4.20\%)} & 1.8507~\textcolor{red}{(-5.44\%)} & 26.75~\textcolor{blue}{(-2.36\%)} & 2.4650~\textcolor{blue}{(+2.35\%)}\\
\midrule
ChatGLM3 (vanilla) &
47.00 & 55.82 & 1.4418 & 29.05 & 2.0541 & 31.84 & 2.2935\\ 
ChatGLM3 (\textsc{TuTri}) &
54.00~\textcolor{red}{(+7.00\%)} & 57.24~\textcolor{red}{(+1.42\%)} & 1.4276~\textcolor{red}{(-1.42\%)} & 39.51~\textcolor{red}{(+10.46\%)} & 1.8642~\textcolor{red}{(-9.50\%)} & 30.62~\textcolor{blue}{(-1.22\%)} & 2.4689~\textcolor{blue}{(+5.85\%)}\\
\bottomrule
\end{tabular}}
\caption{Statistical results of vanilla LLM players VS \textsc{TuTri} players. The decision threshold $p$ for critical decision in ODHR@K and ADR@K is 0.00. Red annotations indicate favorable experimental results, while blue annotations indicate unfavorable experimental results.}
\label{exp_result_table_3}
\end{center}
\end{table*}

\begin{table*}[ht]
\begin{center}
\scalebox{0.64}{
\begin{tabular}{@{}l|l|ll|ll|ll@{}}
\toprule
LLM
& WR ($\uparrow$)& ODHR@2 ($\uparrow$)& DAR@2 ($\downarrow$)& ODHR@3 ($\uparrow$)& DAR@3 ($\downarrow$)& ODHR@4 ($\uparrow$)& DAR@4($\downarrow$)\\
\midrule[\heavyrulewidth]
Gemini-pro (\textsc{TuTri}) &
56.50 & 53.64 & 1.4636 & 34.13 & 1.9461 & 30.36 & 2.3482 \\
\midrule
Gemini-pro + $\textsc{TuTri}^{\prime}$ &
52.50~\textcolor{red}{(-4.00\%)} & 54.33~\textcolor{blue}{(+0.69\%)} & 1.4567~\textcolor{blue}{(-0.69\%)} & 29.88~\textcolor{red}{(-4.25\%)} & 2.0610~\textcolor{red}{(+5.75\%)} & 31.25~\textcolor{blue}{(+0.89\%)} & 2.4219~\textcolor{red}{(+2.46\%)} \\
Gemini-pro + $\textsc{TuTri}^{\prime \prime}$ &
53.50~\textcolor{red}{(-3.00\%)} & 54.59~\textcolor{blue}{(+0.95\%)} & 1.4541~\textcolor{blue}{(-0.95\%)} & 31.95~\textcolor{red}{(-2.18\%)} & 2.0384~\textcolor{red}{(+4.62\%)} & 27.59~\textcolor{red}{(-2.77\%)} & 2.3824~\textcolor{blue}{(-1.14\%)} \\
\bottomrule
\end{tabular}}
\caption{Statistical results of the ablation study. Where $\textsc{TuTri}^{\prime}$ represents the \textsc{TuTri} player which remove the game history reflection module, and $\textsc{TuTri}^{\prime \prime}$ represents the \textsc{TuTri} player which remove the game strategy reflection module. The decision threshold $p$ for critical decision in ODHR@K and ADR@K is 0.15. Red annotations indicate favorable experimental results, while blue annotations indicate unfavorable experimental results.}
\label{exp_result_table_4}
\end{center}
\end{table*}

From the Table \ref{exp_result_table_1}, we can find that (1) Except for ChatGLM3, the WR of other vanilla LLM players and RL players are all above 50.00\%; (2) The performance of GPT-4 is the best, and GPT-4 performs excellently on the 7 evaluation metrics. Especially, the WR of GPT-4 is 63.20\%, 13.20\% higher than 50.00\%.

\subsection{5-players UNO Arena with 5 LLMs}
\label{subsection: 4.2}
To find the best LLM, we place 5 LLMs in a 5-players UNO Arena to compete against each other. We fix the initial playing order of UNO Arena in the sequence of GPT-3.5, GPT-4, Gemini-Pro, Llama 2, and ChatGLM3. We conduct experiment on 200 decks generated randomly. All players are the vanilla LLM players. The results are shown in Table \ref{exp_result_table_2}. 

From the Table \ref{exp_result_table_2}, we can find that (1) GPT-4 has the best performance, with a WR of 24.20\%, 4.2\% higher than the average (20.00\%) and 1.4\% higher than the second highest ranked GPT-3.5. Not only that, GPT-4 also performs the best in other 6 evaluation metrics; (2) ChatGLM3 has the worst performance, with a WR of 15.60\%, which is 4.4\% lower than the average (20.00\%) and 8.6\% lower than the highest ranked GPT-4. Not only that, ChatGLM3 also performs the worst in ODHR@2, DAR@2, DAR@3, ODHR@4, and DAR@4.

\begin{figure}
\vspace{-25pt}
\centering
\includegraphics[width=0.5\textwidth]{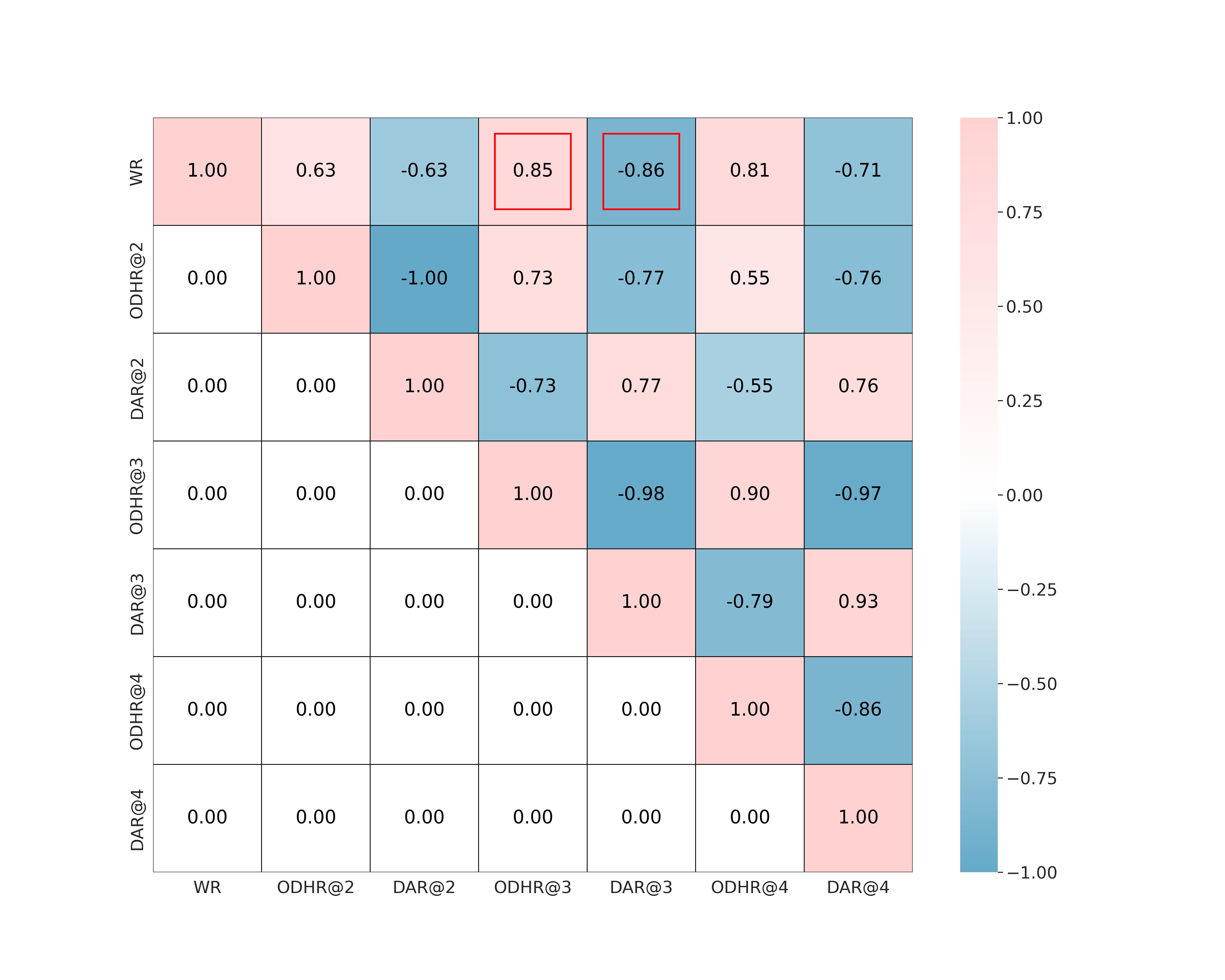}
\caption{The Pearson Correlation Heatmap among WR, ODHR@K (K=2,3,4), and ADR@K (K=2,3,4).}
\label{heat_map}
\vspace{-10pt}
\end{figure}

\begin{figure*}[htbp]
\centering
\includegraphics[width=1\textwidth, height=0.4\textwidth]{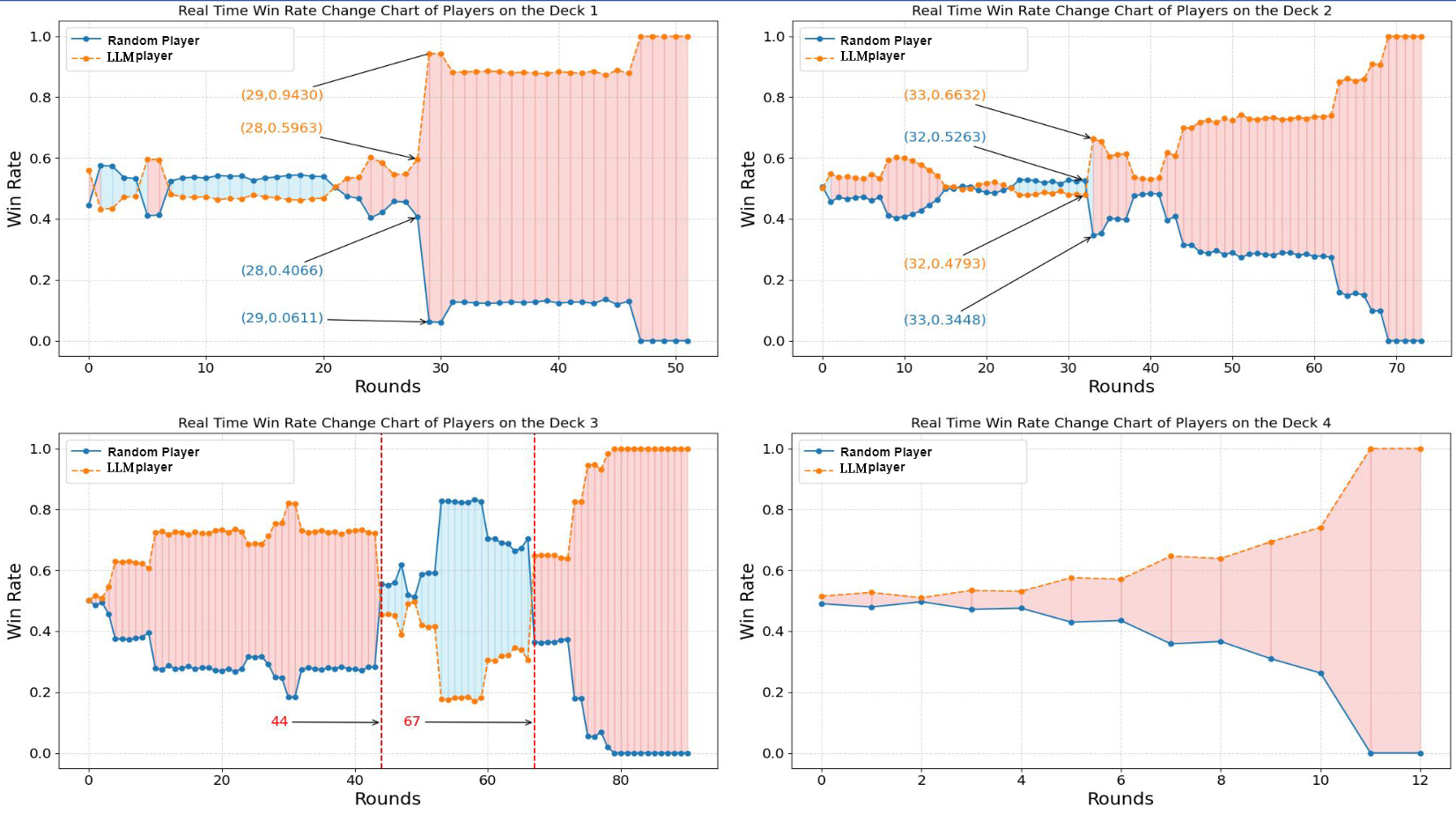}
\caption{GPT-4 (the vanilla LLM player) real-time winning rate variations on 4 decks.}
\label{casestudy}
\end{figure*}

\subsection{Validation of the superiority of the \textsc{TuTri} player compared to the vanilla LLM player} 
\label{subsection: 4.3}
To verify that our \textsc{TuTri} player can improve the sequential decision-making ability of LLMs, we compare the vanilla LLM players (baseline) with \textsc{TuTri} players. We let 5 LLMs serve as the backend LLMs for both the vanilla LLM players and \textsc{TuTri} players, and play two-players UNO Arena on 200 decks generated randomly. The results are shown in Table \ref{exp_result_table_3}. 

From the Table \ref{exp_result_table_3}, we can find that:  (1) All LLMs (the \textsc{TuTri} player) are better than LLM (the vanilla LLM player) on WR, ODHR@2, and DAR@3. Gemini-Pro (the \textsc{TuTri} player) has a 12.50\% higher than Gemini-Pro (the vanilla LLM player) on WR; (2) For ODHR@3, except for Gemini-Pro which performed slightly worse (-2.91\%), the other 4 LLMs achieved good results. For ODHR@4 and DAR@4, GPT-4 and Gemini-Pro both performed well. It can be seen that the \textsc{TuTri} player based on reflection can significantly improve its abilities of sequential decision-making after two rounds of reflection on the summary of game history and the game strategies. The experimental results strongly support the superiority of our \textsc{TuTri} player based reflection over the vanilla LLM player.

\subsection{Ablation studies on \textsc{TuTri} player} 

\label{subsection: 4.4}
To illustrate the necessity of the two reflection modules in the \textsc{TuTri} player, we conduct ablation study. We remove the game history reflection module and the game strategy reflection module from the \textsc{TuTri} players, and conduct two-players UNO Arena with vanilla LLM player respectively. The results are shown in Table \ref{exp_result_table_4}.

From the Table \ref{exp_result_table_4}, we can find that:  (1) After removing the game history reflection module, the WR of decreased by 4\%, the ODHR@3 of decreased 4.25\%, and the DAR@3 of increase by 5.75\%. (2) After removing the game strategy reflection module, the WR of decreased by 3\%, the ODHR@3 of decreased 2.18\%. and the DAR@3 of increase by 4.62\%. The game history holds significant potential information for incomplete information games. Therefore, removing the game history reflection module has a greater adverse impact on the \textsc{TuTri} player.

\section{Discussion}

\subsection{Further Exploration of ODHR@K and ADR@K}
To better analyze the relationship between our unique evaluation metrics (ODHR@K and ADR@K), and the evaluation metric WR, we conduct a Pearson correlation analysis of the experimental results from the Table \ref{exp_result_table_1}. The results are shown in Figure \ref{heat_map}. From the Figure \ref{heat_map}, we can find that (1) WR shows a positive correlation with ODHR@K (K=2,3,4), and simultaneously, WR shows a negative correlation with ADR@K (K=2,3,4); (2) The strongest positive correlation, reaching 0.85, exists between WR and ODHR@3, while the strongest negative correlation, reaching -0.86, exists between WR and DAR@3. Overall, our unique ODHR@K and ADR@K have a good correlation with WR, so they can serve as reference evaluation metrics for evaluating LLMs in UNO Arena.

\subsection{Case Study}
In order to more intuitively see the advantages of LLM versus random player, we conduct a case study. We utilized GPT-4 as the backend LLM for the vanilla LLM player to engage in the game across 4 decks generated randomly, with the random player plays first. We recorded all decision points (for both the vanilla LLM player and the random player) and employed the Monte Carlo method to calculate the real-time percentage change in winning rate for both sides following each decision point. The results are shown in Figure \ref{casestudy}.

From the Figure \ref{casestudy}, we can find that: (1) In UNO Arena, winning rates fluctuate significantly. For example, in deck 1, from round 28 to 29, the random player's winning rate dropped by 34.5\%, while the vanilla LLM player by 34.67\%; (2) Turning points, like rounds 44 and 67 in deck 3, show shifts in dominance. Initially, the vanilla LLM player leads until round 44, then loses advantage until round 67, before regaining control; (3) Brief game durations occur, notably in deck 4, where the agent player consistently makes exceptional decisions, steadily increasing its winning rate until achieving victory. These findings underscore LLM's adeptness at identifying crucial decision junctures and exploiting its capabilities, highlighting its potential in sequential decision-making scenarios.

\section{Conclusion}
In conclusion, LLMs possess the capability for sequential decision-making, as evidenced by the experimental results of LLMs playing the UNO game. Our proposed UNO Arena and unique evaluation metrics enable LLMs to compete with each other in the same UNO Arena game, thereby providing a better dynamic assessment of LLMs' sequential decision-making abilities. Furthermore, we propose that the \textsc{TuTri} player effectively addresses how to enhance LLMs' sequential decision-making abilities for better performance in playing UNO Arena.

\section*{Limitations}
The method of dynamically evaluating the sequential decision-making ability of LLMs using the UNO Arena, as well as the \textsc{TuTri} player, is only applicable to LLMs that support \textit{chat}. The unique evaluation metrics, ODHR@K and ADR@K, introduced in this paper are only applicable to games or tasks with a limited action space.

\bibliography{latex/acl_latex}

\begin{thebibliography}{50}
\expandafter\ifx\csname natexlab\endcsname\relax\def\natexlab#1{#1}\fi

\bibitem[{Achiam et~al.(2023)Achiam, Adler, Agarwal, Ahmad, Akkaya, Aleman, Almeida, Altenschmidt, Altman, Anadkat et~al.}]{achiam2023gpt}
Josh Achiam, Steven Adler, Sandhini Agarwal, Lama Ahmad, Ilge Akkaya, Florencia~Leoni Aleman, Diogo Almeida, Janko Altenschmidt, Sam Altman, Shyamal Anadkat, et~al. 2023.
\newblock Gpt-4 technical report.
\newblock \emph{arXiv preprint arXiv:2303.08774}.

\bibitem[{Aiyappa et~al.(2023)Aiyappa, An, Kwak, and Ahn}]{aiyappa2023trust}
Rachith Aiyappa, Jisun An, Haewoon Kwak, and Yong-Yeol Ahn. 2023.
\newblock \href {http://arxiv.org/abs/2303.12767} {Can we trust the evaluation on chatgpt?}

\bibitem[{Akata et~al.(2023)Akata, Schulz, Coda-Forno, Oh, Bethge, and Schulz}]{akata2023playing}
Elif Akata, Lion Schulz, Julian Coda-Forno, Seong~Joon Oh, Matthias Bethge, and Eric Schulz. 2023.
\newblock Playing repeated games with large language models.
\newblock \emph{arXiv preprint arXiv:2305.16867}.

\bibitem[{Amir(2014)}]{amir2014reasoning}
Eyal Amir. 2014.
\newblock Reasoning and decision making.
\newblock \emph{The Cambridge handbook of artificial intelligence}, pages 191--212.

\bibitem[{Amiri et~al.(2020)Amiri, Shirazi, and Zhang}]{amiri2020learning}
Saeid Amiri, Mohammad~Shokrolah Shirazi, and Shiqi Zhang. 2020.
\newblock Learning and reasoning for robot sequential decision making under uncertainty.
\newblock In \emph{Proceedings of the AAAI Conference on Artificial Intelligence}, volume~34, pages 2726--2733.

\bibitem[{Bouton(1901)}]{bouton1901nim}
Charles~L Bouton. 1901.
\newblock Nim, a game with a complete mathematical theory.
\newblock \emph{The Annals of Mathematics}, 3(1/4):35--39.

\bibitem[{Brookins and DeBacker(2023)}]{brookins2023playing}
Philip Brookins and Jason~Matthew DeBacker. 2023.
\newblock Playing games with gpt: What can we learn about a large language model from canonical strategic games?
\newblock \emph{Available at SSRN 4493398}.

\bibitem[{Chang et~al.(2023)Chang, Wang, Wang, Wu, Yang, Zhu, Chen, Yi, Wang, Wang et~al.}]{chang2023survey}
Yupeng Chang, Xu~Wang, Jindong Wang, Yuan Wu, Linyi Yang, Kaijie Zhu, Hao Chen, Xiaoyuan Yi, Cunxiang Wang, Yidong Wang, et~al. 2023.
\newblock A survey on evaluation of large language models.
\newblock \emph{ACM Transactions on Intelligent Systems and Technology}.

\bibitem[{Chen et~al.(2023)Chen, Wang, Dong, Du, Yan, Yang, Li, Zhao, Qin, Rajmohan et~al.}]{chen2023introspective}
Liting Chen, Lu~Wang, Hang Dong, Yali Du, Jie Yan, Fangkai Yang, Shuang Li, Pu~Zhao, Si~Qin, Saravan Rajmohan, et~al. 2023.
\newblock Introspective tips: Large language model for in-context decision making.
\newblock \emph{arXiv preprint arXiv:2305.11598}.

\bibitem[{Demaine et~al.(2014)Demaine, Demaine, Harvey, Uehara, Uno, and Uno}]{demaine2014uno}
Erik~D Demaine, Martin~L Demaine, Nicholas~JA Harvey, Ryuhei Uehara, Takeaki Uno, and Yushi Uno. 2014.
\newblock Uno is hard, even for a single player.
\newblock \emph{Theoretical Computer Science}, 521:51--61.

\bibitem[{Du et~al.(2021)Du, Qian, Liu, Ding, Qiu, Yang, and Tang}]{du2021glm}
Zhengxiao Du, Yujie Qian, Xiao Liu, Ming Ding, Jiezhong Qiu, Zhilin Yang, and Jie Tang. 2021.
\newblock Glm: General language model pretraining with autoregressive blank infilling.
\newblock \emph{arXiv preprint arXiv:2103.10360}.

\bibitem[{Frankish and Ramsey(2014)}]{frankish2014cambridge}
Keith Frankish and William~M Ramsey. 2014.
\newblock \emph{The Cambridge handbook of artificial intelligence}.
\newblock Cambridge University Press.

\bibitem[{Ganguli et~al.(2023)Ganguli, Schiefer, Favaro, and Clark}]{ganguli2023challenges}
Deep Ganguli, Nicholas Schiefer, Marina Favaro, and Jack Clark. 2023.
\newblock \href {https://www.anthropic.com/index/evaluating-ai-systems} {Challenges in evaluating {AI} systems}.

\bibitem[{Gong et~al.(2023)Gong, Huang, Ma, Vo, Durante, Noda, Zheng, Zhu, Terzopoulos, Fei-Fei et~al.}]{gong2023mindagent}
Ran Gong, Qiuyuan Huang, Xiaojian Ma, Hoi Vo, Zane Durante, Yusuke Noda, Zilong Zheng, Song-Chun Zhu, Demetri Terzopoulos, Li~Fei-Fei, et~al. 2023.
\newblock Mindagent: Emergent gaming interaction.
\newblock \emph{arXiv preprint arXiv:2309.09971}.

\bibitem[{Guo et~al.(2023)Guo, Yang, Yoo, Lin, Iwasawa, and Matsuo}]{guo2023suspicionagent}
Jiaxian Guo, Bo~Yang, Paul Yoo, Bill~Yuchen Lin, Yusuke Iwasawa, and Yutaka Matsuo. 2023.
\newblock \href {http://arxiv.org/abs/2309.17277} {Suspicion-agent: Playing imperfect information games with theory of mind aware gpt-4}.

\bibitem[{Guo et~al.(2024{\natexlab{a}})Guo, Bu, Wang, Ren, Sui, Shang, and Lu}]{guo2024economics}
Shangmin Guo, Haoran Bu, Haochuan Wang, Yi~Ren, Dianbo Sui, Yuming Shang, and Siting Lu. 2024{\natexlab{a}}.
\newblock Economics arena for large language models.
\newblock \emph{arXiv preprint arXiv:2401.01735}.

\bibitem[{Guo et~al.(2024{\natexlab{b}})Guo, Zhang, Ma, Zhang, and Feng}]{guo2024sillm}
Shoutao Guo, Shaolei Zhang, Zhengrui Ma, Min Zhang, and Yang Feng. 2024{\natexlab{b}}.
\newblock Sillm: Large language models for simultaneous machine translation.
\newblock \emph{arXiv preprint arXiv:2402.13036}.

\bibitem[{He et~al.(2022)He, Wang, McAuley, and Majumder}]{he2022controlling}
Zexue He, Yu~Wang, Julian McAuley, and Bodhisattwa~Prasad Majumder. 2022.
\newblock Controlling bias exposure for fair interpretable predictions.
\newblock \emph{arXiv preprint arXiv:2210.07455}.

\bibitem[{Hendrycks et~al.(2021)Hendrycks, Burns, Basart, Zou, Mazeika, Song, and Steinhardt}]{hendrycks2021measuring}
Dan Hendrycks, Collin Burns, Steven Basart, Andy Zou, Mantas Mazeika, Dawn Song, and Jacob Steinhardt. 2021.
\newblock \href {http://arxiv.org/abs/2009.03300} {Measuring massive multitask language understanding}.

\bibitem[{Kroese et~al.(2014)Kroese, Brereton, Taimre, and Botev}]{kroese2014monte}
Dirk~P Kroese, Tim Brereton, Thomas Taimre, and Zdravko~I Botev. 2014.
\newblock Why the monte carlo method is so important today.
\newblock \emph{Wiley Interdisciplinary Reviews: Computational Statistics}, 6(6):386--392.

\bibitem[{Li et~al.(2020)Li, Koyamada, Ye, Liu, Wang, Yang, Zhao, Qin, Liu, and Hon}]{li2020suphx}
Junjie Li, Sotetsu Koyamada, Qiwei Ye, Guoqing Liu, Chao Wang, Ruihan Yang, Li~Zhao, Tao Qin, Tie-Yan Liu, and Hsiao-Wuen Hon. 2020.
\newblock Suphx: Mastering mahjong with deep reinforcement learning.
\newblock \emph{arXiv preprint arXiv:2003.13590}.

\bibitem[{Light et~al.(2023)Light, Cai, Shen, and Hu}]{light2023avalonbench}
Jonathan Light, Min Cai, Sheng Shen, and Ziniu Hu. 2023.
\newblock \href {http://arxiv.org/abs/2310.05036} {Avalonbench: Evaluating llms playing the game of avalon}.

\bibitem[{Littman(1996)}]{littman1996algorithms}
Michael~Lederman Littman. 1996.
\newblock \emph{Algorithms for sequential decision-making}.
\newblock Brown University.

\bibitem[{Madaan et~al.(2023)Madaan, Tandon, Gupta, Hallinan, Gao, Wiegreffe, Alon, Dziri, Prabhumoye, Yang et~al.}]{madaan2023self}
Aman Madaan, Niket Tandon, Prakhar Gupta, Skyler Hallinan, Luyu Gao, Sarah Wiegreffe, Uri Alon, Nouha Dziri, Shrimai Prabhumoye, Yiming Yang, et~al. 2023.
\newblock Self-refine: Iterative refinement with self-feedback.
\newblock \emph{arXiv preprint arXiv:2303.17651}.

\bibitem[{McMullen(2015)}]{mcmullen2015entrepreneurial}
Jeffery~S McMullen. 2015.
\newblock Entrepreneurial judgment as empathic accuracy: A sequential decision-making approach to entrepreneurial action.
\newblock \emph{Journal of Institutional Economics}, 11(3):651--681.

\bibitem[{Mnih et~al.(2013)Mnih, Kavukcuoglu, Silver, Graves, Antonoglou, Wierstra, and Riedmiller}]{mnih2013playing}
Volodymyr Mnih, Koray Kavukcuoglu, David Silver, Alex Graves, Ioannis Antonoglou, Daan Wierstra, and Martin Riedmiller. 2013.
\newblock Playing atari with deep reinforcement learning.
\newblock \emph{arXiv preprint arXiv:1312.5602}.

\bibitem[{Mooney(1997)}]{mooney1997monte}
Christopher~Z Mooney. 1997.
\newblock \emph{Monte carlo simulation}.
\newblock 116. Sage.

\bibitem[{OpenAI(2022)}]{ChatGPT}
OpenAI. 2022.
\newblock Introducing chatgpt.
\newblock \url{https://openai.com/blog/chatgpt}.
\newblock Accessed: 2023-09-30.

\bibitem[{Pfann(2021)}]{TacklUNO2021}
Bernhard Pfann. 2021.
\newblock Tackling the uno card game with reinforcement learning.
\newblock \emph{Towards Data Science: tackling-uno-card-game-with-reinforcement-learning}.

\bibitem[{Sainz et~al.(2023)Sainz, Campos, Garc{\'\i}a-Ferrero, Etxaniz, de~Lacalle, and Agirre}]{sainz-etal-2023-nlp}
Oscar Sainz, Jon Campos, Iker Garc{\'\i}a-Ferrero, Julen Etxaniz, Oier~Lopez de~Lacalle, and Eneko Agirre. 2023.
\newblock \href {https://doi.org/10.18653/v1/2023.findings-emnlp.722} {{NLP} evaluation in trouble: On the need to measure {LLM} data contamination for each benchmark}.
\newblock In \emph{Findings of the Association for Computational Linguistics: EMNLP 2023}, pages 10776--10787, Singapore. Association for Computational Linguistics.

\bibitem[{Shinn et~al.(2024)Shinn, Cassano, Gopinath, Narasimhan, and Yao}]{shinn2024reflexion}
Noah Shinn, Federico Cassano, Ashwin Gopinath, Karthik Narasimhan, and Shunyu Yao. 2024.
\newblock Reflexion: Language agents with verbal reinforcement learning.
\newblock \emph{Advances in Neural Information Processing Systems}, 36.

\bibitem[{Silver et~al.(2017)Silver, Schrittwieser, Simonyan, Antonoglou, Huang, Guez, Hubert, Baker, Lai, Bolton et~al.}]{silver2017mastering}
David Silver, Julian Schrittwieser, Karen Simonyan, Ioannis Antonoglou, Aja Huang, Arthur Guez, Thomas Hubert, Lucas Baker, Matthew Lai, Adrian Bolton, et~al. 2017.
\newblock Mastering the game of go without human knowledge.
\newblock \emph{nature}, 550(7676):354--359.

\bibitem[{Taneja et~al.(2024)Taneja, Laird, Yan, Musuvathi, and Lahiri}]{taneja2024llm}
Jubi Taneja, Avery Laird, Cong Yan, Madan Musuvathi, and Shuvendu~K Lahiri. 2024.
\newblock Llm-vectorizer: Llm-based verified loop vectorizer.
\newblock \emph{arXiv preprint arXiv:2406.04693}.

\bibitem[{Team et~al.(2023)Team, Anil, Borgeaud, Wu, Alayrac, Yu, Soricut, Schalkwyk, Dai, Hauth et~al.}]{team2023gemini}
Gemini Team, Rohan Anil, Sebastian Borgeaud, Yonghui Wu, Jean-Baptiste Alayrac, Jiahui Yu, Radu Soricut, Johan Schalkwyk, Andrew~M Dai, Anja Hauth, et~al. 2023.
\newblock Gemini: a family of highly capable multimodal models.
\newblock \emph{arXiv preprint arXiv:2312.11805}.

\bibitem[{Touvron et~al.(2023)Touvron, Lavril, Izacard, Martinet, Lachaux, Lacroix, Rozi{\`e}re, Goyal, Hambro, Azhar et~al.}]{touvron2023llama}
Hugo Touvron, Thibaut Lavril, Gautier Izacard, Xavier Martinet, Marie-Anne Lachaux, Timoth{\'e}e Lacroix, Baptiste Rozi{\`e}re, Naman Goyal, Eric Hambro, Faisal Azhar, et~al. 2023.
\newblock Llama: Open and efficient foundation language models.
\newblock \emph{arXiv preprint arXiv:2302.13971}.

\bibitem[{Wang et~al.(2023{\natexlab{a}})Wang, Zhao, Ouyang, Wang, and Shen}]{wang2023chatcad}
Sheng Wang, Zihao Zhao, Xi~Ouyang, Qian Wang, and Dinggang Shen. 2023{\natexlab{a}}.
\newblock Chatcad: Interactive computer-aided diagnosis on medical image using large language models.
\newblock \emph{arXiv preprint arXiv:2302.07257}.

\bibitem[{Wang et~al.(2023{\natexlab{b}})Wang, Liu, Zheng, Qi, Chen, Yang, Zhao, Wang, Song, and Huang}]{wang2023avalons}
Shenzhi Wang, Chang Liu, Zilong Zheng, Siyuan Qi, Shuo Chen, Qisen Yang, Andrew Zhao, Chaofei Wang, Shiji Song, and Gao Huang. 2023{\natexlab{b}}.
\newblock \href {http://arxiv.org/abs/2310.01320} {Avalon's game of thoughts: Battle against deception through recursive contemplation}.

\bibitem[{Wang et~al.(2023{\natexlab{c}})Wang, Liu, Zhang, Yao, Heinecke, and Yu}]{wang2023drdt}
Yu~Wang, Zhiwei Liu, Jianguo Zhang, Weiran Yao, Shelby Heinecke, and Philip~S Yu. 2023{\natexlab{c}}.
\newblock Drdt: Dynamic reflection with divergent thinking for llm-based sequential recommendation.
\newblock \emph{arXiv preprint arXiv:2312.11336}.

\bibitem[{Wei et~al.(2022)Wei, Tay, Bommasani, Raffel, Zoph, Borgeaud, Yogatama, Bosma, Zhou, Metzler et~al.}]{wei2022emergent}
Jason Wei, Yi~Tay, Rishi Bommasani, Colin Raffel, Barret Zoph, Sebastian Borgeaud, Dani Yogatama, Maarten Bosma, Denny Zhou, Donald Metzler, et~al. 2022.
\newblock Emergent abilities of large language models.
\newblock \emph{arXiv preprint arXiv:2206.07682}.

\bibitem[{Wythoff(1907)}]{wythoff1907modification}
Willem~A Wythoff. 1907.
\newblock A modification of the game of nim.
\newblock \emph{Nieuw Arch. Wisk}, 7(2):199--202.

\bibitem[{Xi et~al.(2023)Xi, Chen, Guo, He, Ding, Hong, Zhang, Wang, Jin, Zhou et~al.}]{xi2023rise}
Zhiheng Xi, Wenxiang Chen, Xin Guo, Wei He, Yiwen Ding, Boyang Hong, Ming Zhang, Junzhe Wang, Senjie Jin, Enyu Zhou, et~al. 2023.
\newblock The rise and potential of large language model based agents: A survey.
\newblock \emph{arXiv preprint arXiv:2309.07864}.

\bibitem[{Xu et~al.(2024)Xu, Guan, Greene, and Kechadi}]{xu2024benchmark}
Cheng Xu, Shuhao Guan, Derek Greene, and M-Tahar Kechadi. 2024.
\newblock \href {http://arxiv.org/abs/2406.04244} {Benchmark data contamination of large language models: A survey}.

\bibitem[{Xu et~al.(2023)Xu, Wang, Li, Luo, Wang, Liu, and Liu}]{xu2023exploring}
Yuzhuang Xu, Shuo Wang, Peng Li, Fuwen Luo, Xiaolong Wang, Weidong Liu, and Yang Liu. 2023.
\newblock \href {http://arxiv.org/abs/2309.04658} {Exploring large language models for communication games: An empirical study on werewolf}.

\bibitem[{You et~al.(2019)You, Li, Guo, Wang, and Lu}]{you2019combinational}
Yang You, Liangwei Li, Baisong Guo, Weiming Wang, and Cewu Lu. 2019.
\newblock Combinational q-learning for dou di zhu.
\newblock \emph{arXiv preprint arXiv:1901.08925}.

\bibitem[{Zeng et~al.(2022)Zeng, Liu, Du, Wang, Lai, Ding, Yang, Xu, Zheng, Xia et~al.}]{zeng2022glm}
Aohan Zeng, Xiao Liu, Zhengxiao Du, Zihan Wang, Hanyu Lai, Ming Ding, Zhuoyi Yang, Yifan Xu, Wendi Zheng, Xiao Xia, et~al. 2022.
\newblock Glm-130b: An open bilingual pre-trained model.
\newblock \emph{arXiv preprint arXiv:2210.02414}.

\bibitem[{Zeng et~al.(2024)Zeng, Chen, Liu, Jiang, and Jia}]{zeng2024mrgsm8k}
Zhongshen Zeng, Pengguang Chen, Shu Liu, Haiyun Jiang, and Jiaya Jia. 2024.
\newblock \href {http://arxiv.org/abs/2312.17080} {Mr-gsm8k: A meta-reasoning revolution in large language model evaluation}.

\bibitem[{Zha et~al.(2019)Zha, Lai, Cao, Huang, Wei, Guo, and Hu}]{zha2019rlcard}
Daochen Zha, Kwei-Herng Lai, Yuanpu Cao, Songyi Huang, Ruzhe Wei, Junyu Guo, and Xia Hu. 2019.
\newblock Rlcard: A toolkit for reinforcement learning in card games.
\newblock \emph{arXiv preprint arXiv:1910.04376}.

\bibitem[{Zhai et~al.(2024)Zhai, Bai, Lin, Pan, Tong, Zhou, Suhr, Xie, LeCun, Ma, and Levine}]{zhai2024finetuning}
Yuexiang Zhai, Hao Bai, Zipeng Lin, Jiayi Pan, Shengbang Tong, Yifei Zhou, Alane Suhr, Saining Xie, Yann LeCun, Yi~Ma, and Sergey Levine. 2024.
\newblock \href {http://arxiv.org/abs/2405.10292} {Fine-tuning large vision-language models as decision-making agents via reinforcement learning}.

\bibitem[{Zhang et~al.(2023)Zhang, Xu, and Deng}]{zhang2023exploring}
Jintian Zhang, Xin Xu, and Shumin Deng. 2023.
\newblock Exploring collaboration mechanisms for llm agents: A social psychology view.
\newblock \emph{arXiv preprint arXiv:2310.02124}.

\bibitem[{Zhou et~al.(2023)Zhou, Zhu, Chen, Chen, Zhao, Chen, Lin, Wen, and Han}]{zhou2023dont}
Kun Zhou, Yutao Zhu, Zhipeng Chen, Wentong Chen, Wayne~Xin Zhao, Xu~Chen, Yankai Lin, Ji-Rong Wen, and Jiawei Han. 2023.
\newblock \href {http://arxiv.org/abs/2311.01964} {Don't make your llm an evaluation benchmark cheater}.

\end{thebibliography}

\appendix
\large{\textbf{Appendix}}

\section{UNO Game}
\label{UNO_appendix}
In this section of appendix, you will learn what UNO Game is, and in order to facilitate the evaluation of LLMs with the UNO Game, we have made some slight modifications to it.
\subsection{Game Objective}
In the modified UNO game, We simply set the game objective as to be the first player to clear out the hand. Players play alternatively(2 players) or in circle manner(3 or more players) and strive to achieve the unique goal. It should be noted that if the cards in the deck are exhausted by players, the player with the fewest number of cards in hand wins, so there may be multiple winners in the same game.
\subsection{UNO Cards}
UNO comprises 3 categories of cards: number cards, function cards, and wild cards. In total, UNO features 108 cards.
\begin{itemize}
  \item \textbf{The Number Cards}: the number cards can be expressed in the form of COLOR + NUMBER, where COLOR is belong to the set $\{Red, Blue, Yellow, Green\}$, and NUMBER is an integer from $0$ to $9$. It is important to note that there is only one 0-number card per color, while there are two 1-9 number cards per color. There are a total of 76 number cards.
  \item \textbf{The Function Cards}: the function cards can be expressed in the form of COLOR + FUNCTION, where COLOR is belong to the set $\{Red, Blue, Yellow, Green\}$, and FUNCTION is belong to the set $\{Skip, Reverse, Draw Two\}$. There are two cards of the same COLOR for each FUNCTION. There are a total of 24 function cards.
      \begin{itemize}
          \item \textbf{Skip}: the player's next player skips this round of play.
          \item \textbf{Reverse}: the player reverses the order of play (from clockwise to counterclockwise, or from counterclockwise to clockwise).
          \item \textbf{Draw Two}: The player's next player draws two cards and skips this round of play.
      \end{itemize}
  \item \textbf{The Wild Cards}: the wild card includes 4 Black Wild cards and 4 Black Wild Draw Four cards, totaling 8 cards.
        \begin{itemize}
          \item \textbf{Wild}: the player selects one color from the COLOR set $\{Red, Blue, Yellow, Green\}$ as the new color for the top card in the discard pile.
          \item \textbf{Wild Draw Four}: the player selects one color from the COLOR set $\{Red, Blue, Yellow, Green\}$ as the new color for the top card in the discard pile, and the player's next player draws 4 cards.
      \end{itemize}
\end{itemize}

\subsection{Game Progress}
First, deal each player 7 initial cards in clockwise order, then continue drawing cards until a number card is drawn and set as the top card of the initial discard pile. All players take rounds playing cards in clockwise order(it will be reversed by a reverse card) until a player runs out of his cards or the draw pile is exhausted, signaling the end of the game.

\subsection{Legal Decision(Action)}
In every round of the game, player in charge can using rules to match the top card of the discard pile otherwise pick up a new card into hand. The rules, or say, the legal decisions consists of several sorts:
\begin{itemize}
    \item \textbf{Draw Card}: If a player does not have any cards to play during their playing round, they must draw a card, or their previous player used Draw Two or Wild Draw Four cards to make the player draw multiple cards.
    \item \textbf{Select Card}: In a player's playing round, they need to play a card that matches either the COLOR, NUMBER, or FUNCTION of the top card in the discard pile, or play a Wild card(include Wild Draw Four card) to match. The card played by the player then becomes the new top card.
    \item \textbf{Select Color}: After selecting either Wild Card or Wild Draw Four Card, the player needs to convert the color of the current top card to one of $\{Red, Blue, Yellow, Green\}$.
    \item \textbf{Select ChallengeFlag}: The use of the Wild Draw Four card may be illegal. After a player plays a Wild Draw Four card, their next player can choose to challenge its use. If the player who played the Wild Draw Four card still holds non-Wild cards matching the color of the current top card, the use of the Wild Draw Four card is illegal. Possible scenarios are as follows: (1) If the player's play is illegal and their next player challenges it, the player must draw 4 cards, and their next player faces no penalty; (2) If the player's play is legal and their next player challenges it, the player's next player must draw 6 cards. (3) If the player's next player does not challenge, regardless of the legality of the player's play, the player's next player must draw 4 cards.
Note that Challenge is not a stand-alone action to complete turns, it should be accompanied by a card draw or card match action.
\end{itemize}

\section{Prompt}
\label{prompt}
Here is the prompt design for the entire experiment.

\subsection{Select Card}
\begin{figure*}[htbp]
\centering
\includegraphics[width=1\textwidth]{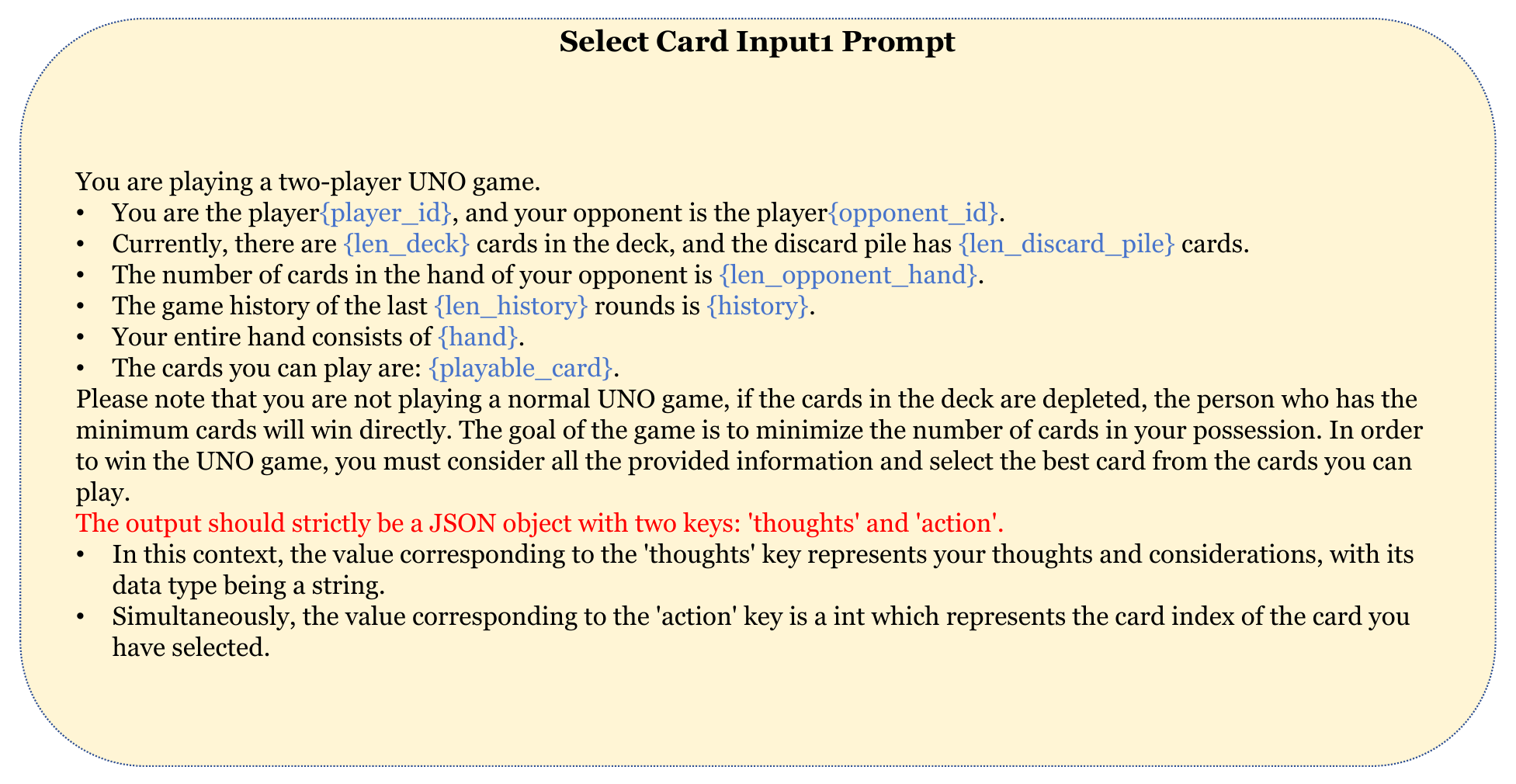}
\caption{The input1 prompt of the select card shared by the vanilla LLM player and the \textsc{TuTri} player.}
\label{figure-5}
\end{figure*}

\begin{figure*}[htbp]
\centering
\includegraphics[width=1\textwidth]{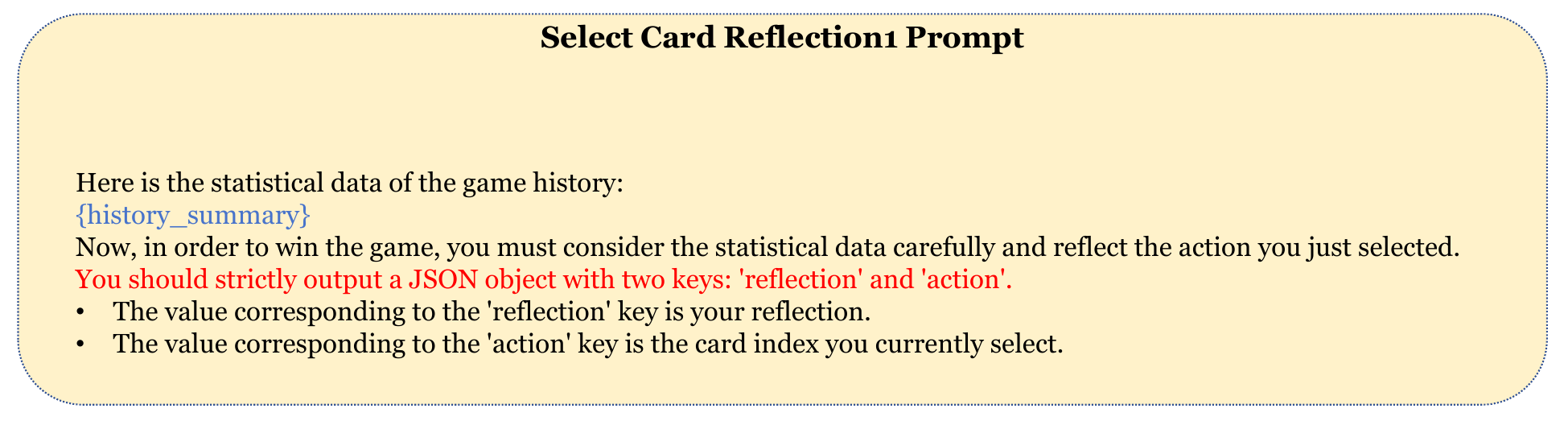}
\caption{The game history reflection module prompt of the select card for \textsc{TuTri} player.}
\label{figure-6}
\end{figure*}

\begin{figure*}[htbp]
\centering
\includegraphics[width=1\textwidth]{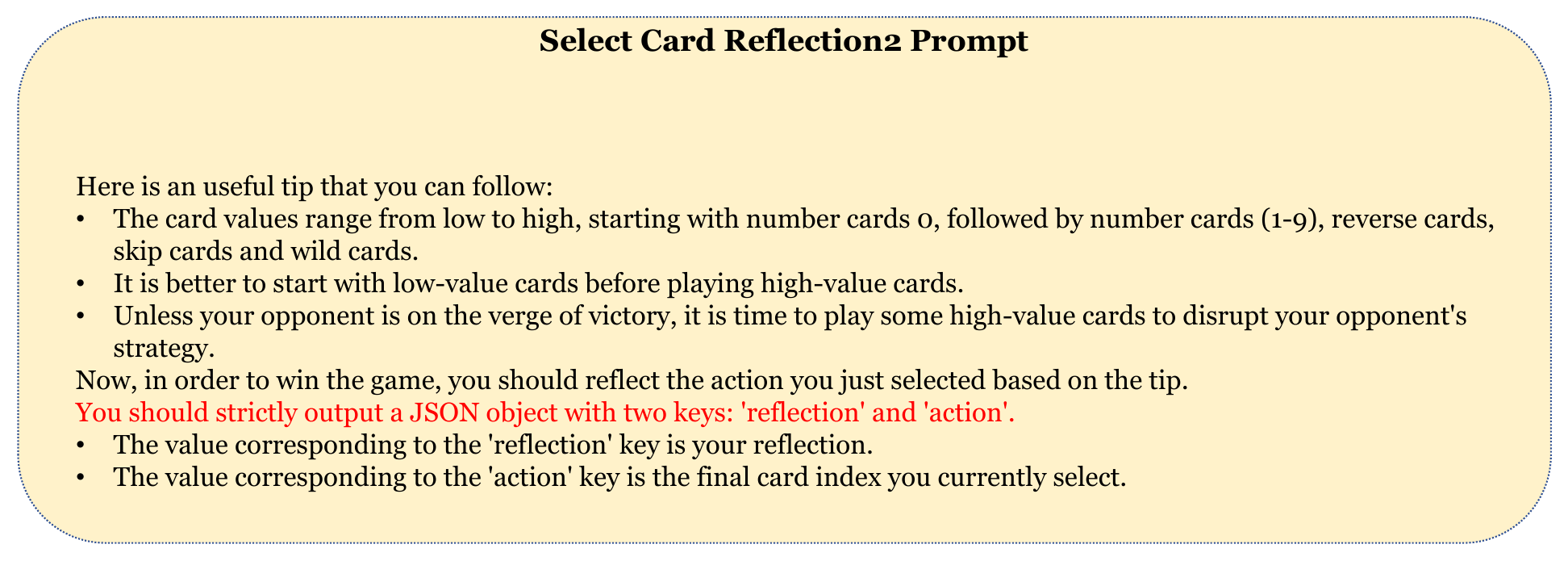}
\caption{The game strategy reflection module prompt of the select card for \textsc{TuTri} player.}
\label{figure-7}
\end{figure*}

The input1 prompt of the select card shared by the vanilla LLM player and the \textsc{TuTri} player is shown in the Figure \ref{figure-5}. The game history Rreflection module prompt of the select card for the \textsc{TuTri} player is shown in the Figure \ref{figure-6}. The game strategy reflection module prompt of the select card for the \textsc{TuTri} player is shown in the Fugure \ref{figure-7}.

\subsection{Select Color}
\begin{figure*}[htbp]
\centering
\includegraphics[width=1\textwidth]{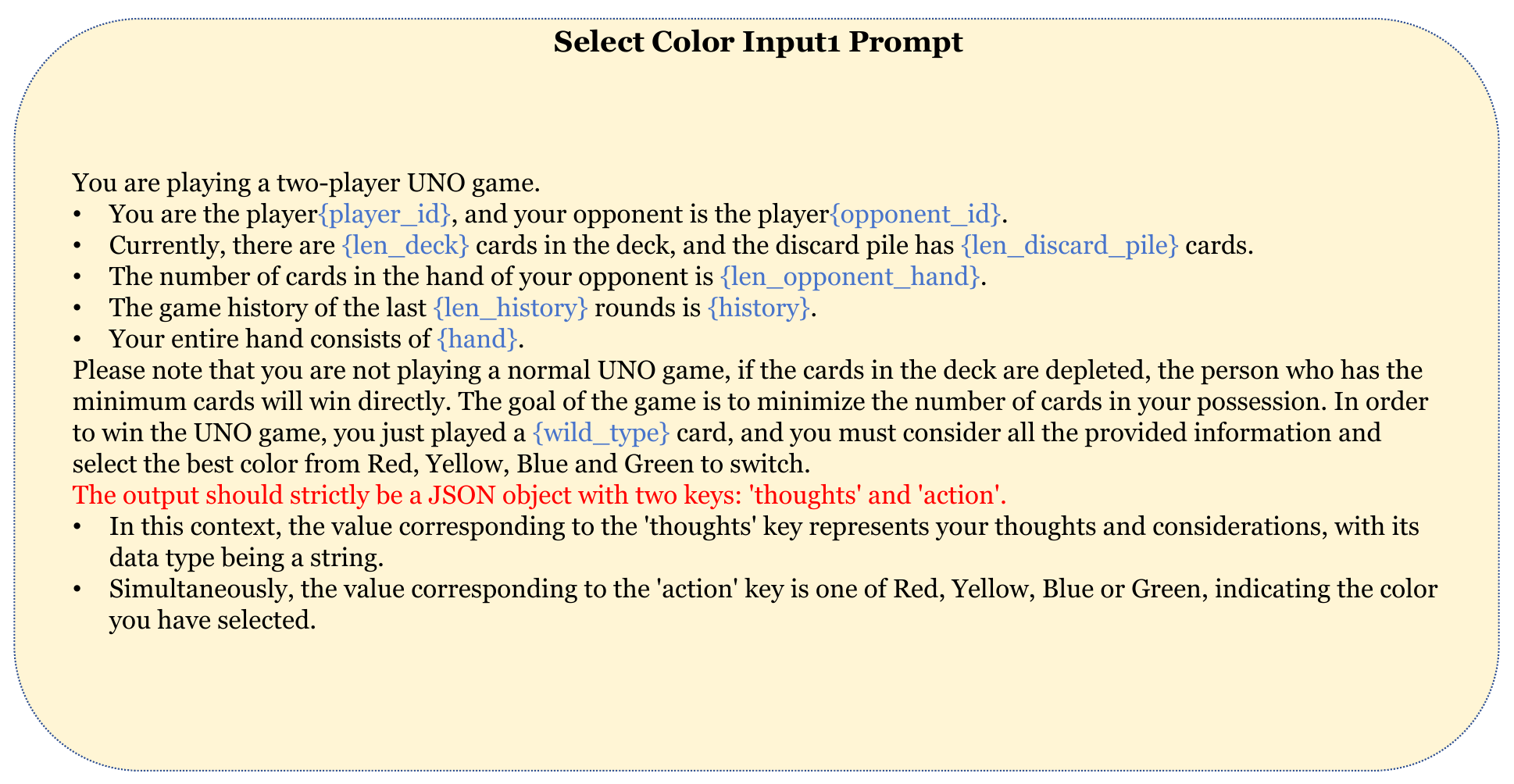}
\caption{The input1 prompt of the select color shared by the vanilla LLM player and the \textsc{TuTri} player.}
\label{figure-8}
\end{figure*}

\begin{figure*}[htbp]
\centering
\includegraphics[width=1\textwidth]{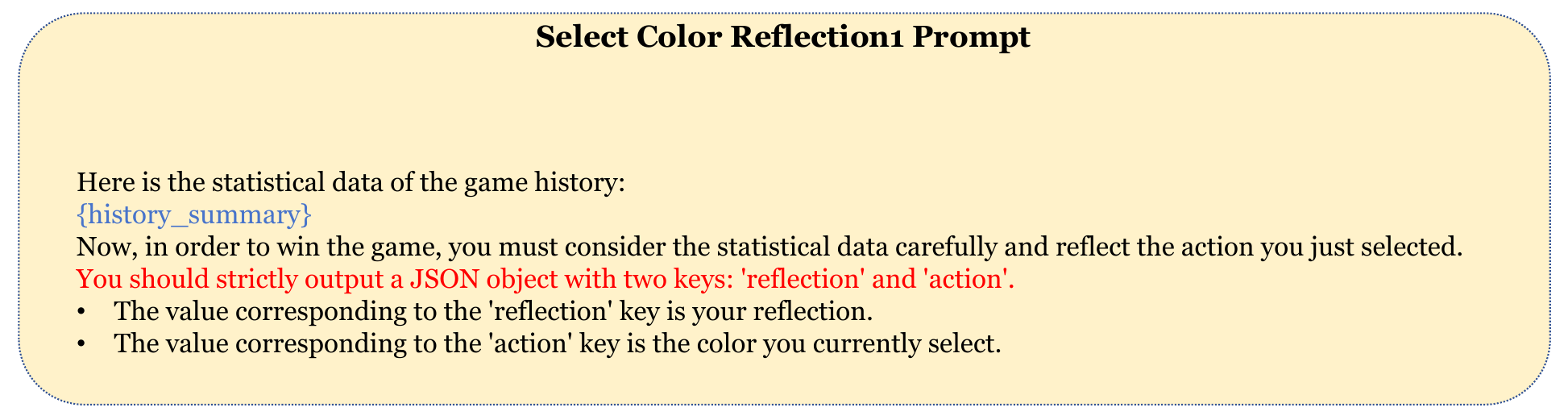}
\caption{The game history reflection module prompt of the select color for \textsc{TuTri} player.}
\label{figure-9}
\end{figure*}

\begin{figure*}[htbp]
\centering
\includegraphics[width=1\textwidth]{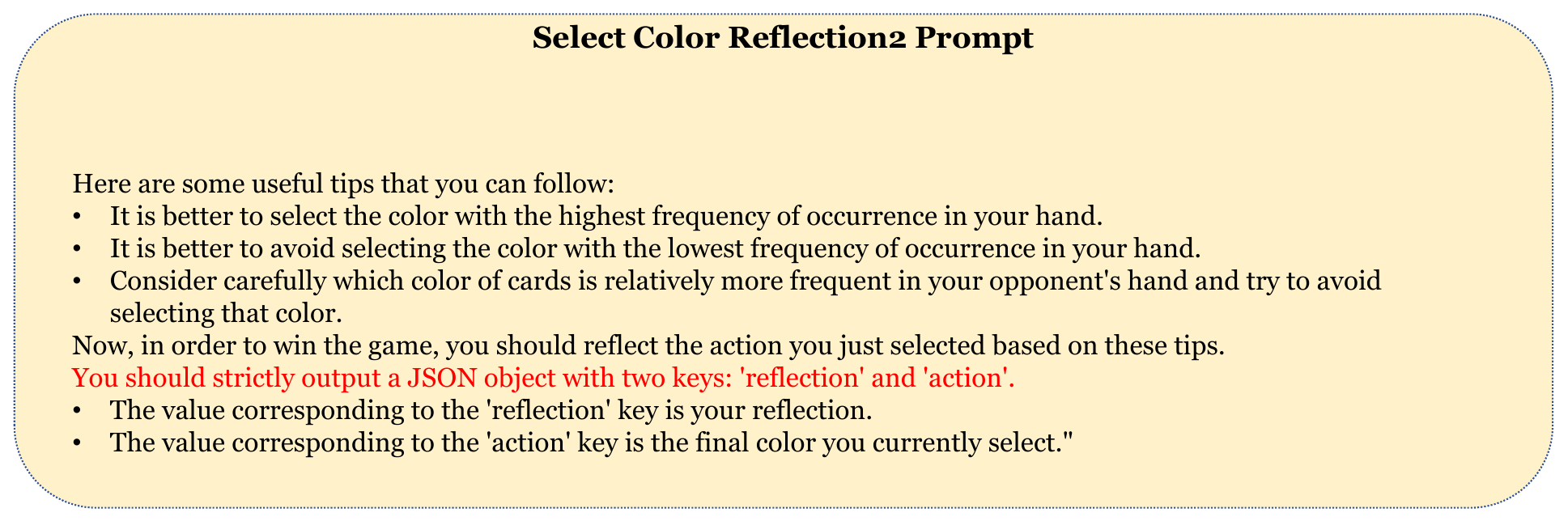}
\caption{The game strategy reflection module prompt of the select color for \textsc{TuTri} player.}
\label{figure-10}
\end{figure*}

The input1 prompt of the select color shared by the vanilla LLM player and the \textsc{TuTri} player is shown in the Figure \ref{figure-8}. The game history reflection module prompt of the select color for the \textsc{TuTri} player is shown in the Figure \ref{figure-9}. The game strategy reflection module prompt of the select color for the \textsc{TuTri} player is shown in the Fugure \ref{figure-10}.

\subsection{Select ChallengeFlag}
\begin{figure*}[htbp]
\centering
\includegraphics[width=1\textwidth]{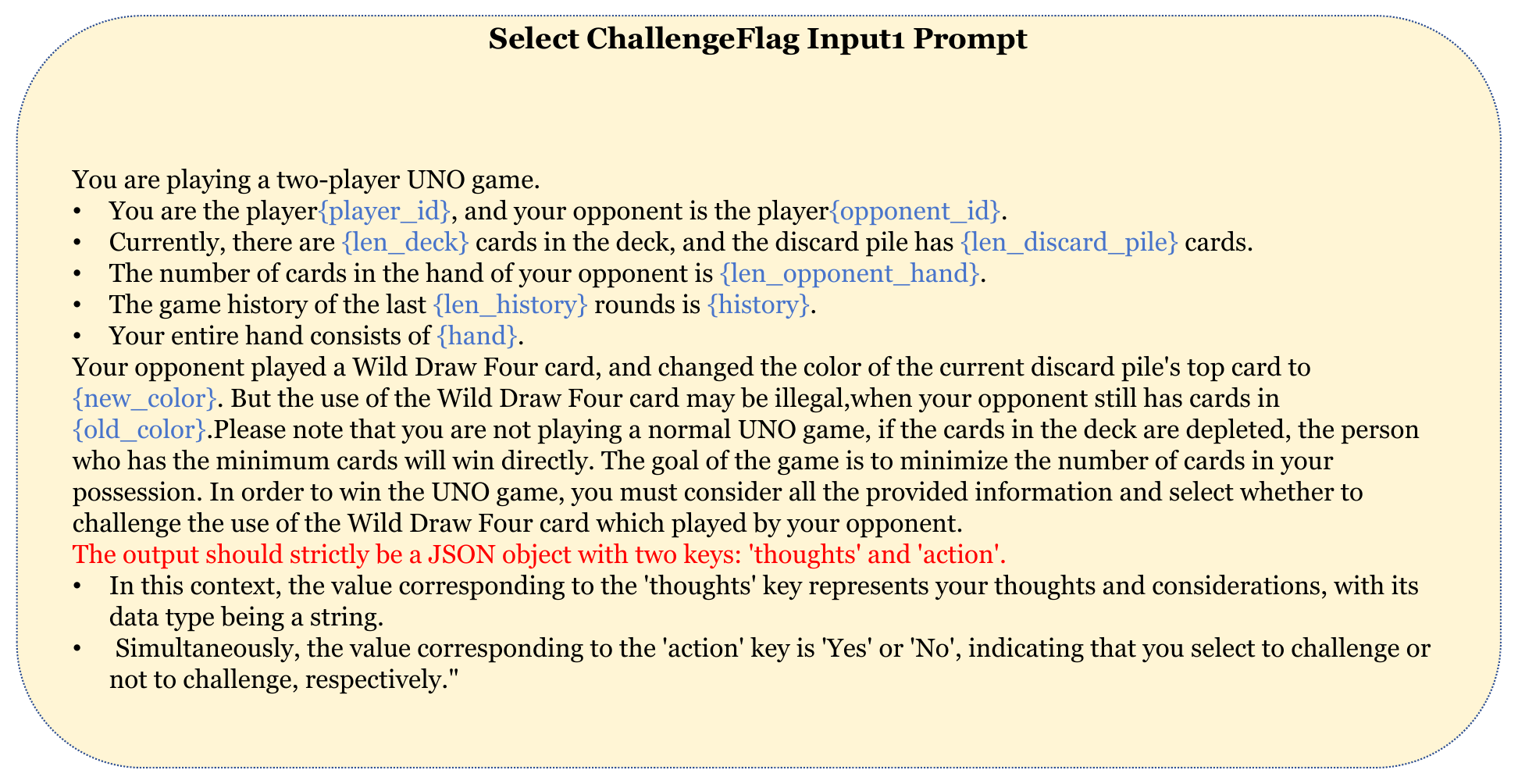}
\caption{The input1 prompt of the select challengeFlag shared by the vanilla LLM player and the \textsc{TuTri} player.}
\label{figure-11}
\end{figure*}

\begin{figure*}[htbp]
\centering
\includegraphics[width=1\textwidth]{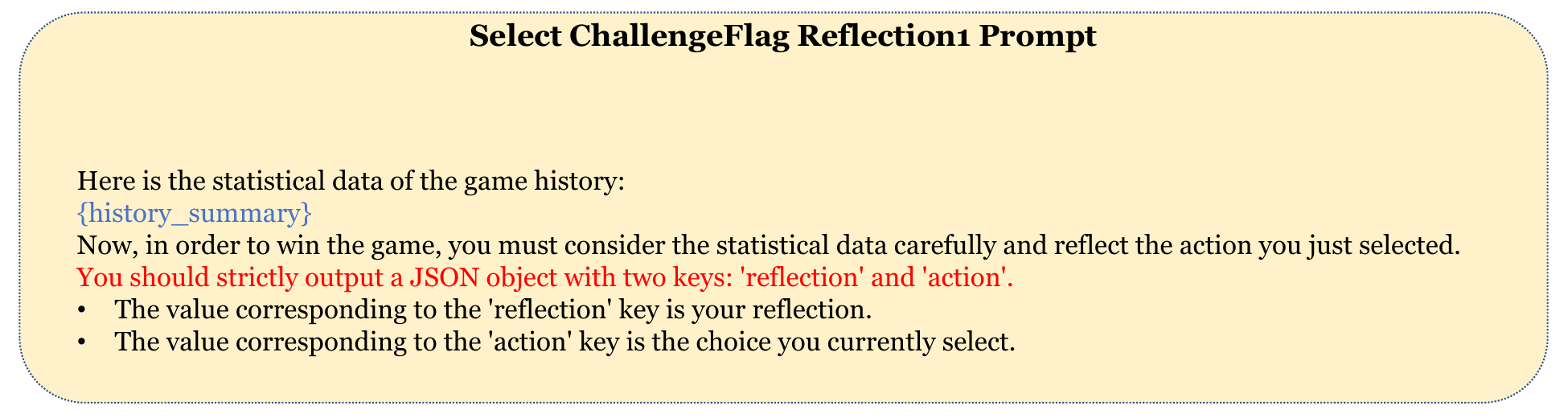}
\caption{The game history reflection module prompt of the select challengeFlag for \textsc{TuTri} player.}
\label{figure-12}
\end{figure*}

\begin{figure*}[htbp]
\centering
\includegraphics[width=1\textwidth]{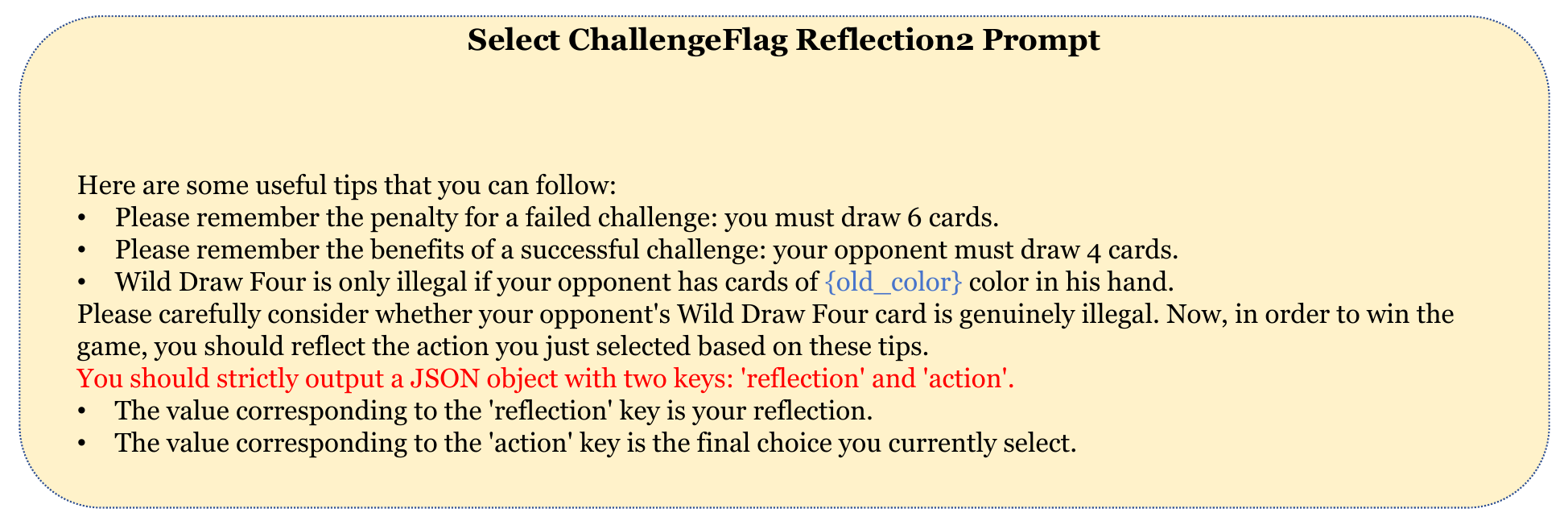}
\caption{The game strategy reflection module prompt of the select challengeFlag for \textsc{TuTri} player.}
\label{figure-13}
\end{figure*}

The input1 prompt of the select challengeFlag shared by the vanilla LLM player and the \textsc{TuTri} player is shown in the Figure \ref{figure-11}. The game history reflection module prompt of the select challengeFlag for the \textsc{TuTri} player is shown in the Figure \ref{figure-12}. The game strategy reflection module prompt of the select challengeFlag for the \textsc{TuTri} player is shown in the Fugure \ref{figure-13}.

\end{document}